\documentclass[nohyperref]{article}

\usepackage{microtype}
\usepackage{graphicx}
\usepackage{booktabs} 
\usepackage{hyperref}
\usepackage{url}
\usepackage{algpseudocode}

\usepackage{amsmath,amsfonts,bm}

\def\eqref#1{equation~\ref{#1}}
\def\Eqref#1{Eq.~(\ref{#1})}

\def\1{\bm{1}}

\DeclareMathAlphabet{\mathsfit}{\encodingdefault}{\sfdefault}{m}{sl}
\SetMathAlphabet{\mathsfit}{bold}{\encodingdefault}{\sfdefault}{bx}{n}

\usepackage[table]{xcolor}

\usepackage{multirow}
\usepackage{graphicx}
\usepackage[utf8]{inputenc} 
\usepackage[T1]{fontenc}    
\usepackage{hyperref}       
\usepackage{url}            
\usepackage{booktabs}       
\usepackage{amsfonts}       
\usepackage{nicefrac}       
\usepackage{microtype}      
\usepackage{amsmath}
\usepackage{dsfont}

\usepackage{amssymb}

\usepackage{multicol}
\usepackage{subcaption}
\usepackage[font=small,labelfont=bf]{caption}

\usepackage[mathscr]{euscript}

\newcommand{\beq}{\begin{equation}}
\newcommand{\eeq}{\end{equation}}
\def\be {\begin{equation}}
\def\ee {\end{equation}}
\def\bs#1\es{\begin{split}#1\end{split}}
\def\ba#1\ea{\begin{align}#1\end{align}}
\def\baed#1\eaed{\begin{aligned}#1\end{aligned}}
\def\bged#1\eged{\begin{gathered}#1\end{gathered}}
\def\bea{\begin{eqnarray}}
\def\eea{\end{eqnarray}}

\usepackage{hyperref}

\usepackage[accepted]{icml2022}

\usepackage{amsmath}
\usepackage{amssymb}
\usepackage{mathtools}
\usepackage{amsthm}

\usepackage[capitalize,noabbrev]{cleveref}

\theoremstyle{plain}
\newtheorem{theorem}{Theorem}[section]

\newtheorem{lemma}[theorem]{Lemma}

\theoremstyle{definition}
\newtheorem{definition}[theorem]{Definition}

\theoremstyle{remark}

\usepackage[textsize=tiny]{todonotes}

\icmltitlerunning{Koopman Q-learning}

\begin{document}

\twocolumn[
\icmltitle{Koopman Q-learning: Offline Reinforcement Learning\\via Symmetries of Dynamics}



\icmlsetsymbol{equal}{*}

\begin{icmlauthorlist}

\icmlauthor{Matthias Weissenbacher}{RI}
\icmlauthor{Samarth Sinha}{VI}
\icmlauthor{Animesh Garg}{VI}
\icmlauthor{Yoshinobu Kawahara }{RI,KU}

\end{icmlauthorlist}

\icmlaffiliation{RI}{RIKEN Center for Advanced Intelligence Project, Japan}
\icmlaffiliation{VI}{ Vector Institute, University of Toronto, Canada}
\icmlaffiliation{KU}{Institute of Mathematics for Industry, Kyushu University, Japan }

\icmlcorrespondingauthor{Matthias Weissenbacher}{matthias.weissenbacher@a.riken.jp}

\icmlkeywords{Machine Learning, ICML}

\vskip 0.3in
]



\printAffiliationsAndNotice{}  

\begin{abstract}
  Offline reinforcement learning  leverages large datasets to train policies without interactions with the environment. The learned policies may then be deployed in real-world settings where interactions are costly or dangerous.
  Current algorithms over-fit to the training dataset and as a consequence perform poorly when deployed to out-of-distribution generalizations of the environment. 
  We aim to address these limitations by learning a Koopman latent representation which allows us to infer symmetries of the system's underlying dynamic.
The latter is then utilized to extend the otherwise static offline dataset during training; 
 this constitutes a novel data augmentation framework which reflects the system's dynamic and is thus to be interpreted as an exploration of the environment's state space.  
  To obtain the symmetries we employ Koopman theory in which non-linear dynamics are represented in terms of a linear operator acting on the space of measurement functions of the system.  We provide novel theoretical results on the existence and nature of symmetries relevant for control systems such as reinforcement learning settings.
     Moreover, we empirically evaluate our method on several benchmark offline reinforcement learning tasks and datasets including D4RL, Metaworld and Robosuite and find that by using our framework we consistently improve the state-of-the-art of model-free Q-learning methods.
\end{abstract}

\section{Introduction}
The recent impressive advances in reinforcement learning (RL) range from robotics,  to strategy games  and recommendation systems \citep{RL0,recom}.  Reinforcement learning is canonically regarded as an active learning process - also referred to as online RL - where the agent interacts with the environment at each training run. 
In contrast,  offline RL algorithms learn from large, previously collected static datasets, and thus do not rely on environment interactions \citep{agarwal2020optimistic,OfRL1,OfRL2}.
Online data collection is performed by simulations or by means of real world interactions e.g.\,robotics and in either scenario interactions maybe costly and/or dangerous.

In principle offline datasets only need to be collected once which alleviates the before-mentioned shortcomings of costly online interactions. 
Offline datasets are typically collected using behavioral policies for the specific task ranging from, random policies, or near-optimal policies to human demonstrations. 
In particular, being able to leverage the latter is a major advantage of offline RL over online approaches, and then the learned policies can be deployed or fine-tuned on the desired environment.   
Offline RL has successfully been applied to learn agents that outperform the behavioral policy used to collect the data \citep{CQL, BRAC, agarwal2020optimistic, OfRL1}. 
However algorithms admit major shortcomings in regard to over-fitting and overestimating the true state-action values of the distribution. 
One solution was recently proposed by \citet{S4RL}, where they tested several data augmentation schemes to improve the performance
and generalization capabilities of the learned policies.

\begin{figure*}[t!]
\vspace{0.20 cm}
\begin{center}
\includegraphics[width=\textwidth]{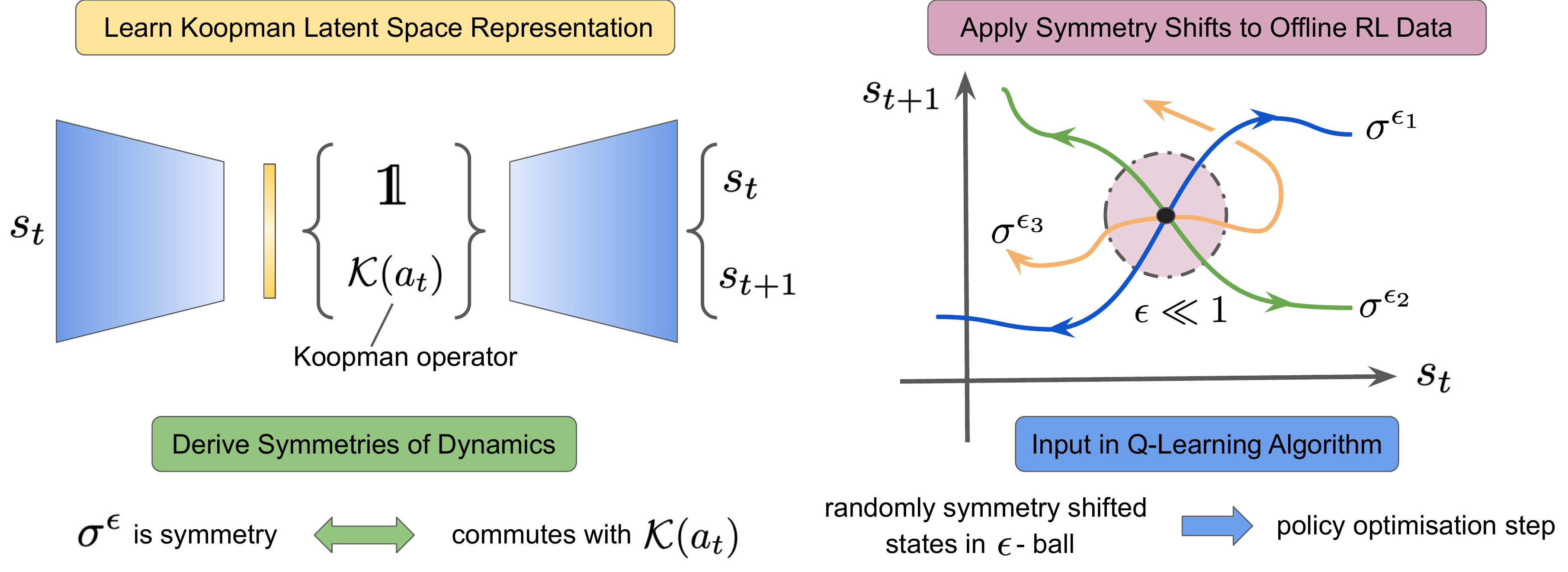}
\end{center}
\caption{Overview of Koopman Q-learning. The states of the data point $(s_t,s_{t+1},a_t)$ are shifted along symmetry trajectories parametrized by $\epsilon_{1,2,3}$ for constant $a_t$. Symmetry transformations can be combined to reach other specific subset of the $\epsilon$-ball region. }
\label{fig:main1}
\vspace{-5pt}
\end{figure*}

However, despite the recent progress, learning from  offline demonstrations is a tedious endeavour as the  dataset typically does not cover the full state-action space. 
Moreover,  offline RL algorithms per definition do not admit the possibility for further environment exploration to refine their distributions towards an optimal policy.
It was argued previously that is basically impossible for an offline RL agent to learn an optimal policy as the generalization to near data generically leads to compounding errors such as overestimation bias \citep{CQL}.
In this paper, we look at offline RL through the lens of Koopman spectral theory in which non-linear dynamics are represented in terms of a linear operator acting on the space of measurement functions of the system. 
Through which the representation the symmetries of the dynamics may be inferred directly, and can then 
be used to guide data augmentation strategies see Figure \ref{fig:main1}.  
We further provide theoretical results on the existence on nature of symmetries relevant for control systems such as reinforcement learning.
More specifically, we apply Koopman spectral theory by:
first learning symmetries of the system's underlying dynamic in a self-supervised fashion from the static dataset, and
second employing the latter to extend the offline dataset at training time by out-of-distribution values. 
As this reflects the system's dynamics the additional data is to be interpreted as an exploration of the environment's state space. \vspace{0.1cm}

Some prior works have explored symmetry of the state-action space in the context of Markov Decision Processes (MDP's) \citep{higgins,homo,vanderPol} since many control tasks exhibit apparent symmetries e.g.\,the cart-pole 
is translation symmetric w.r.t.\,its position.
The paradigm introduced in this work is of a different nature entirely. 
The distinction is twofold: first, the symmetries are learned in a self-supervised way and are in general not apparent to the developer;
second: we concern with symmetry transformation from state tuples $(s_t,s_{t+1}) \to (\tilde s_t, \tilde s_{t+1})$  which leave the action invariant. The latter, are inferred from the dynamics inherited by the behavioral policy of the underlying offline data. 
In other words we seek to derive a neighbourhood around a MDP tuple in the offline dataset in which the behavioral policy is likely to choose the same action based on its dynamics in the environment. 
In practice the Koopman latent space representation is learned in a self-supervised manner by training to predict the next state using a VAE model \citep{vae}.

To summarize, in this paper, we propose Koopman Forward (Conservative) Q-learning (KFC): a model-free 
Q-learning algorithm which uses the symmetries in the dynamics of the environment to guide data 
augmentation strategies.
We also provide thorough theoretical justifications for KFC.
Finally, we empirically test our approach on several challenging benchmark datasets from D4RL \citep{d4rl}, 
MetaWorld 
\citep{metaworld} and Robosuite \citep{robosuite} and find that by using KFC we can improve the 
state-of-the-art on most benchmark offline reinforcement learning tasks.

\section{Preliminaries and background}
\label{prelim}

\subsection{Offline RL \& Conservative Q-learning}

Reinforcement learning algorithms train policies to maximize the cumulative reward received by an agent who interacts with an environment. Formally the setting is given by a Markov decision process $ (\mathcal{S}, \mathcal{A}, \rho, r, \gamma)$, with state space $\mathcal{S}$, action space $\mathcal{A}$, and  $\rho(s_{t+1}|s_t, a_t)$  the transition density function from the current state and action to the next state. Moreover,  $\gamma$ is the discount factor and $r(S_t)$ the reward function. 
At any discrete time the agent chooses an action $a_t \in \mathcal{A}$ according to its underlying  policy $\pi_\theta (a_t | s_t) $ based on the information of the current state $s_t \in \mathcal{S}$ where the policy is parametrized by $\theta$.

We focus on  the Actor-Critic methods for continuous control tasks in the following.
 In deep RL the parameters $\theta$ are the weights in a deep neural network function approximation of the policy or Actor as well as the state-action value function $Q$ or  Critic, respectively, and are optimized  by gradient decent.
The agent i.e.\,the Actor-Critic is trained  to maximize the expected $\gamma$-discounted cumulative reward
$ \mathbb{E}_{\pi} [ \sum_{t=0}^T \gamma^ t \, r_\pi(s_t,a_t) ]$, with respect to the policy network i.e.\,its parameters  $\theta$. For notational simplicity we omit the explicit dependency of the latter in the remainder of this work.
 Furthermore the state-action value function  $Q(s_t, a_t)$, returns  the value of performing a given action $a_t$ while being in the state $s_t$. The Q-function is trained by minimizing the Bellman error  as
  \vspace{0.1 cm}
\beq\label{eq:policystep}
\hat Q_{i+1} \leftarrow \arg \min_{Q} \mathbb{E} \Big[\big( r_t + \gamma \, \hat Q_i(s_{t+1}, a_{t+1}) - Q(s_t, a_t) \big)^2\Big]\,.
\eeq\vspace{0.05 cm}
This  is commonly referred to as the i$^\text{th}$ policy evaluation step where the hat denotes the target Q-function.
In offline RL one aims to learn an optimal policy for the given the dataset $\mathcal{D} = \bigcup_{k=1,\dots,\text{\#data-points}} \big( s_t,a_t,r_t,s_{t+1}\big)_k$ as the option for exploration of the MDP is not available. The policy is optimized to maximize the state-action value function via the policy improvement 
 \beq\label{eq:policy_improvement2}
 \pi_{i+1} \leftarrow \arg \max_{\pi} \mathbb{E}_{s_t \sim \mathcal{D}} \big[\hat Q_i\big(s_t, \pi_{}(a_t|s_t)\big) \big] \;.
 \eeq
\newline
{\bf CQL algorithm:} 
CQL is built on top of a Soft Actor-Critic algorithm (SAC) \citep{SAC}, which employs soft-policy iteration  of a stochastic policy \citep{haarnoja17a}.  A policy entropy regularization term is added to the policy improvement step in \Eqref{eq:policy_improvement2} as
\vspace{0.2 cm}
 \beq\label{eq:policy_improvement3}
 \pi_{i+1} \leftarrow \arg \max_{\pi} \underset{s_t \sim \mathcal{D}}{\mathbb{E}} \Big[\hat Q_i\big(s_t, \pi_{}(a_t|s_t)\big)  - \alpha \log  \pi_{}(a_t|s_t)\Big] \,
 \eeq
where $\alpha$ either is a fixed hyperparameter or may be chosen to  be trainable. 

CQL reduces the overestimation of state-values  - in particular those out-of distribution  from  $\mathcal D$.  It achieves this by regularizing the Q-function  in \Eqref{eq:policystep} by a term minimizing its values over out of distribution randomly sampled actions. In the following $a_{t+1}$ is given by the prediction of the policy $\pi_{i}(a_{t+1}|s_{t+1})$. The policy optimisation step \Eqref{eq:policystep} is modified by a regulizer term  
\beq
\label{eq:policystepCQL}
 \arg \min_{Q} \underset{ s_t \sim \mathcal{D}}{\mathbb{E}} \Big[ \log \sum_a \exp\big( Q(s_{t}, a) \big) - \underset{ a \sim \pi(s_t) }{\mathbb{E}} \big[Q(s_t, a) \big] \Big]\ .
\eeq
Where we have omitted the hyperparameter  $\tilde\alpha$  balancing the regulizer term in \Eqref{eq:policystepCQL}.

\subsection{Koopman theory}
 Historically,  the Koopman theoretic perspective of dynamical systems was introduced to describe the evolution of measurements of Hamiltonian systems \citep{Koo31,Mez05}. 
The underlying dynamic of most modern reinforcement learning tasks is of non-linear nature, i.e.\,the agents actions lead to changes of it state described by a complex non-linear dynamical system.
In contrast to linear systems which are completely characterized by their spectral decomposition non-linear systems lack such a unified characterisation.

 The Koopman operator theoretic framework describes non-linear dynamics via a linear infinite-dimensional Koopman operator and thus inherits certain tools applicable to linear control systems \citep{MMS20,KKB21}.  In practice one aims to find a finite-dimensional  representation of the Koopman operator which is equivalent to obtaining a coordinate transformations in which the non-linear dynamics are approximately linear.
A general non-affine control system is governed by the system of non-linear ordinary differential equations (ODEs) as \vspace{-0.05cm}
\beq \label{eq:non_affine}
\dot{s} = f(s, a) \;\; ,
\eeq
where $s,$ is the n-dimensional state vector, $a$ the m-dimensional action vector with $ (s,a) \in \mathcal{S} \times \mathcal{A} $ the state-action-space. Moreover, $\dot{s} = \partial_t s$ is the time derivative, and $f$ is some general non-linear  - at least $\mathcal{C}^ 1$-differentiable - vector valued function. For a discrete time system, \Eqref{eq:non_affine} takes the form \vspace{-0.1cm}
\beq\label{eq:nonaffine2}
s_{t+1} = F(s_t, a_t) \;\; ,
\eeq
where $s_t$ denotes the state at time $t$ where F is at least $\mathcal{C}^1$-differentiable vector valued function. 

\vspace{0.2cm}
\begin{definition}[Koopman operator]
Let $\mathscr{K}(\mathcal{S} \times \mathcal{A} )$ be the (Banach) space of all measurement functions (observables). Then the Koopman operator $\mathcal{K}: \mathscr{K}(\mathcal{S} \times \mathcal{A} ) \to \mathscr{K}(\mathcal{S} \times \mathcal{A} )$ is defined by
\beq
 \mathcal{K} g(s_t,a_t) = g \big( F(s_t,a_t) , a_{t+1}\big) = g(s_{t+1},a_{t+1}) \;, \;\; 
\eeq
for all observables  $g \in \mathscr{K}(\mathcal{S} \times \mathcal{A})$ where $g : \mathcal{S} \times \mathcal{A}  \to \mathbb{R}$.
\end{definition}

Many systems can be modeled by a bilinearisation where the action enters the controlling equations~\ref{eq:non_affine} linearly as
\beq
f(s, a) = f_{0}(s) + \sum_{i}^{m}f_{i}(s) a_i \;\; ,
\eeq 
for  $f_i,\, i=0,\dots,m$ \,$\mathcal{C}^1$-differentiable-vector valued functions.
In that case the action of the Koopman operator takes the simple form \vspace{-0.1cm}
\beq\label{eq:BilinKoop}
g(s_{t+1}) = \mathcal{K}(a) g(s_{t}) =\Big( \mathcal{K}_0 + \sum_{i}^{m} \mathcal{K}_i a_i \Big) g(s_{t}) \;,\vspace{-0.1cm}
\eeq
 $ \forall g \in \mathscr{K}(\mathcal{S})$ where $ \mathscr{K}(\mathcal{S})$ is a  (Banach) space of measurement functions $\mathcal{K}_0 ,\mathcal{K}_i $ decompose the Koopman operator in a the free and forcing-term, respectively. Details on the existence of a representation as in \Eqref{eq:BilinKoop} are discussed in \cite{Goswami2017GlobalBA}.
Associated with a Koopman operator is its eigenspectrum, that is, the eigenvalues $\lambda$, and the corresponding eigenfunctions $\varphi_\lambda (s,a)$, such that
\beq
\mathcal{K} \varphi_\lambda (s,a) = \lambda \varphi_\lambda (s,a)\;\;.
\eeq
In practice one derives a finite set of observable 
$$ \vec{g} = (g_1,\dots,g_N) \;\; ,$$
in which case the approximation to the Koopman operator admits a finite-dimensional matrix representation.  The $N\times N$ matrix representing the Koopman operator  may  be diagonalized by a matrix $U$ containing the eigen-vectors of $\mathcal{K}$ as columns. In which case the eigen-functions are derived by $\vec{\varphi} = U \vec{g}$ and one infers from \Eqref{eq:non_affine} that
$
\dot{\varphi}_i = \lambda_i \varphi_i   \;, \;\; \text{for} \;\; i=1,\dots,N $
with eigenvalues $\lambda_{i=1,\dots,N}$. These ODEs admit  simple solutions for their time-evolution namely the exponential functions $\exp(\lambda_i t)$.

\section{The Koopman forward framework}
\label{KFC_algo}
 We initiate our discussion with a focus on symmetries in dynamical control systems in Subsection~\ref{sec:theory_symmetries} where we additionally  present some theoretical results. We then proceed in Subsection \ref{sec:KFC} by presenting the Koopman forward framework for Q-learning based on CQL.
Moreover, we discuss the algorithm as well as the Koopman Forward model's deep neural network architecture.\newline

\subsection{Symmetries of dynamical control systems}
\label{sec:theory_symmetries}

{\bf Informal Overview:} Our theoretical results culminate in theorem \ref{thm:main} and \ref{thm:main2}, which provides a road-map on how specific symmetries of the dynamical control system are to be inferred from a VAE forward prediction model. 
In particular, theorem \ref{thm:main} guarantees that the procedure leads to  symmetries of the system (at least locally) and theorem \ref{thm:main2} that actual new data points can be derived by applying symmetry transformations to existing ones. To make this topic more approachable we have shifted some technical details to the appendix \ref{app-KTh_and_Sym}. 

For the reader who wants to skip this technical section directly to the algorithm in section \ref{sec:KFC} let us summarize a couple of take-away messages. Trajectories of many RL-environments are often solutions of ODEs. Symmetries can map solutions  i.e. trajectories to each other. Thus having  solutions (offline RL data) the symmetry map provides one the tools to generate novel data points. However, symmetry maps are hard to come by. In this section we provide a principled approach on how to derive this symmetry maps for control system using Koopman theory. Vaguely an operator is a symmetry generator if it commutes with the Koopman operator.

\vspace{0.2 cm}
 In general, a  symmetry group  $\Sigma$ of the state space may be any subgroup of the group of isometrics of the Euclidean space. We  consider Euclidean state spaces  and in particular,  $\Sigma$ invariant compact subsets thereof.
 \begin{definition} [Equivariant Dynamical System]
Consider the dynamical system $\dot{s} = f(s) $ and let $\Sigma$ be a group acting on the state-space $\mathcal{S}$. Then the system is called $\Sigma$-equivariant if $
f (\sigma \cdot s) = \sigma  \cdot f(s) \;, \;\; \text{for} \; s \in \mathcal{S} \;,\;\; \forall \sigma \in \Sigma $.
For a discrete time dynamical system $s_{t+1} = F(s_t)$ one defines equivariance analogously, namely if 
$F (\sigma \cdot s_t) = \sigma  \cdot F(s_t)$ ,  for $\; s_t \in \mathcal{S} \;,\;\; \forall \sigma \in \Sigma $.
\end{definition} 
\vspace{-0.13cm}
 A system of ODEs  may be represented by the set of its solutions which are locally  connected by Lie Groups. In other words the symmetry maps are given by elements of the Lie Group which map between solutions.
 \begin{definition}[Symmetries of ODEs] A symmetry of a system of ODEs on a  locally-smooth structure is a locally-defined diffeomorphism  that maps the set of all solutions to itself.
\end{definition}
 \vspace{-0.03cm}
 \begin{definition} [Local Lie Group]
\label{def:lie}
A parametrized set of transformations $\sigma^\epsilon : \mathcal{S} \to \mathcal{S}$  with $s \mapsto \tilde s(s,\epsilon) $ for $\epsilon \in (\epsilon_{low},\epsilon_{high})$ where $\epsilon_{low} < 0 < \epsilon_{high}$ is a one-parameter local Lie group if\footnote{The points (1) and (2) in definition \ref{def:lie} imply the existence of an inverse element  ${\sigma^\epsilon}^{-1} = \sigma^{-\epsilon}$ for $  |\epsilon| \ll 1 $.  A local Lie group satisfies the group axioms for sufficiently small parameters values;  it may not be a group on the entire set. } \vspace{-0.05cm}
\begin{enumerate} \vspace{-0.1cm}
    \item  $\sigma^0$ is the identity map; for $\epsilon=0$ such that $\tilde s(s, 0) = s$.\vspace{-0.1cm}
    \item  $\sigma^{\epsilon_1}\sigma^{\epsilon_2} =\sigma^{\epsilon_2}\sigma^{\epsilon_1} = \sigma^{\epsilon_1 +\epsilon_2}   $ for every $ |\epsilon_1|, |\epsilon_2| \ll 1 $.
   \vspace{-0.1cm} \item $\tilde s(s,\epsilon)$  admits a Taylor series expansion in $\epsilon$; in a neighbourhood of $s$  determined by $\epsilon=0$ as $
     \tilde s(s,\epsilon) = s + \epsilon\,\zeta(s) + \mathcal{O}(\epsilon^2)
    $.
\end{enumerate}
\end{definition}
The Koopman operator of equivariant dynamical system is reviewed in the appendix \ref{app-KTh_and_Sym}.
Let us next turn to the case relevant for RL, namely control systems. 
In the remainder of this section we focus on dynamical systems given as in \Eqref{eq:nonaffine2} and \Eqref{eq:BilinKoop}.

\begin{definition} [Action-Equivariant Dynamical System]
\label{def:act_equi_DS}
Let $\bar\Sigma$ be a group acting on the state-action-space $\mathcal{S} \times \mathcal{A}$ of a general control system as in \Eqref{eq:nonaffine2} such that it acts as the identity operation on $\mathcal{A}$ i.e.
$
 \sigma\cdot (s_t , a_t) = (\sigma|_{\mathcal{S}} \cdot s_t,a_t) \; ,\;\; \forall \sigma \in \bar\Sigma \;\;.
$
Then the system is called $\bar\Sigma$-action-equivariant if 
\beq
F (\sigma \cdot(s_t,a_t)) = \sigma|_{\mathcal{S}}  \cdot   F (s_t,a_t) \;, \;\; \text{for} \; (s_t,a_t) \in \mathcal{S}\times \mathcal{A} \; .
\eeq 
\end{definition}

\begin{lemma}
\label{lem:barphi}
The map $\bar \ast  : \bar \Sigma \times \mathscr{K}(\mathcal{S} \times \mathcal{A} ) \rightarrow \mathscr{K}(\mathcal{S} \times \mathcal{A} ) $ given by
$
(\sigma \bar\ast g )(s,a) \longmapsto  g( \sigma^{-1} \cdot s,a) $ 
 defines a group action on the Koopman space of observables $\mathscr{K}(\mathcal{S} \times \mathcal{A} )$. 
\end{lemma}
\begin{theorem}
\label{thm:Sym2}
A Koopman operator $\mathcal{K}$  of a $\bar\Sigma$-action-equivariant system $s_{t+1} = F(s_t,a_t)$ satisfies   \vspace{-0.05 cm}
\beq
[\sigma \bar\ast (\mathcal{K} g)](s_t,a_t) = [\mathcal{K} (\sigma \bar\ast g)](s_t,a_t) \;\; .
\eeq 
\end{theorem}
In particular, it is easy to see that the biliniarisation in \Eqref{eq:BilinKoop} is  $\Sigma$-action-equivariant if
$
f_i(\sigma|_{\mathcal{S}} \cdot s) = \sigma|_{\mathcal{S}} \cdot f_i(s)\;,\;\; \forall i=0,\dots,m
$.
Let us thus turn our focus on the  relevant case of a control system    $s_{t+1} = \tilde F (s_t,a_t)$ which admits a Koopman operator description as 
\beq\label{eq:KA}
g(s_{t+1}) = \mathcal{K}(a_t)g(s_{t}) \;,\;\; \text{for}\; a_t\in \mathcal{A} \,, \;\forall g \in \mathscr{K}(\mathcal{S})\;\;,
\eeq
where $\{ \mathcal{K}(a) \}_{a \in \mathcal{A}}$ is a family of operators with analytical dependence on $a \in \mathcal{A}$.
Note that the bilinearisation in \Eqref{eq:BilinKoop} is a special case of \Eqref{eq:KA}.
\footnote{See Appendix~\ref{app-KTh_and_Sym} for the details. We found that, in practice the Koopman operator leaned by the neural nets is diagonalizable almost everywhere. }
\vspace{0.1 cm}
\begin{definition}[$ \hat\Sigma$-Action-Equivariant Control System ] 
\label{def:hatphi}
 We refer to a control system as in \Eqref{eq:KA}  admitting a symmetry map $\hat \ast$  as  $\hat\Sigma$-action-equivariant.
The definition of the map $\hat\ast$ is a bit more involved, see the appendix \ref{lem:hatphiA}. Informally, we note that it will connect a symmetry map depending on the action space $\sigma_a$ and the Koopman observable as
$
(\sigma_a \hat\ast g )(s) \longmapsto g(  \sigma^{-1} \cdot s) 
$.
 The latter, defines a group action on the Koopman space of observables $\mathscr{K}(\mathcal{S})$. 
\end{definition}

\begin{theorem}
\label{thm:main}
Let $s_{t+1} = \tilde F (s_t,a_t)$ be a $\hat\Sigma$-action-equivariant  control system i.e.\,admitting  a symmetry action as in definition \ref{def:hatphi} and  a Koopman operator representation as
\beq\label{eq:goodsys}
g(s_{t+1}) = \mathcal{K}(a_t)g(s_{t}) \;,\;\; \text{for}\; a_t\in \mathcal{A} \,, \;\forall g \in \mathscr{K}(\mathcal{S}) \;\;.
\eeq
Then 
\beq
\label{eq:mainrel}
\big[\sigma_{a_t} \hat\ast \big(\mathcal{K}(a_t) g\big) \big](s_t) =\big[\mathcal{K}(a_t) \, \big(\sigma_{a_t}  \hat\ast g \big)\big](s_t) \;\;.
\eeq
Moreover, a control system obeying equations~\ref{eq:goodsys} and \ref{eq:mainrel} is $\hat\Sigma$-action-equivariant locally if $g^{-1}$ exists for a neighborhood of $s_t$, i.e.\,then $\sigma \cdot \tilde F(s_t,a_t) =  \tilde F( \sigma \cdot (s_t,a_t))$.
\end{theorem}
 To allow an easier transition to the empirical section  we introduce  the notation  $E: \mathcal{S} \to \mathscr{K}(S)$ and $D: \mathscr{K}(S) \to \mathcal{S}$, denoting the $\mathcal{C}^1$-differentiable encoder and decoder to and from the finite-dimensional Koopman space approximation, respectively; \,$E \circ D = D \circ E = id$. 
 Data-points may be shifted by symmetry transformations of solutions of ODEs as discussed next.

\begin{theorem}
\label{thm:main2}
Let $s_{t+1} = \tilde F (s_t,a_t)$ be a control system as in \Eqref{eq:goodsys} and $\sigma_{a_t}$ an operator obeying \Eqref{eq:mainrel}. 
Then $\sigma_{a_t}^\epsilon : \big(s_t,s_{t+1},a_t\big)  \longmapsto  \big(\tilde s_t, \tilde s_{t+1},a_t\big) $ with 
\bea\label{eq:symshift}
\tilde s_t &=& D\Big(\,\big(\mathds{1} \;+\; \epsilon \, \sigma_{a_t}\big)\hat\ast E(s_t) \Big) \; ,\\ 
\nonumber  \tilde s_{t+1} &=& D\Big(\,\big(\mathds{1} \;+\; \epsilon \,\sigma_{a_t} \big) \hat\ast E(s_{t+1}) \Big) \;\; \,
\eea
is a one-parameter local Lie symmetry group of ODEs.
In other words one can use a symmetry transformation to shift both $s_t$ as well as $s_{t+1}$ such that  $\tilde s_{t+1} = \tilde F (\tilde s_t,a_t)$.
\footnote{No assumptions on the Koopman operator are imposed. Moreover, note that the equivalent theorem holds when $\epsilon \,\sigma_{a_t} \to \sum_{I=1}\epsilon_I \,\sigma^I_{a_t}$ to be a local N-parameter Lie group. }  
\end{theorem}
\subsection{KFC algorithm}
\label{sec:KFC}
The previous section provides a theoretical foundation  to derive symmetries of dynamical control systems based on a Koopman latent space representation.   A vague take-away message is that an operator is a symmetry generator if it commutes with the Koopman operator.  Thus the algorithmic challenge of our framework is twofold. Firstly, to find a numerical approximation of the Koopman space and Koopman operator and secondly to derive operators which commute with the latter.

In practice the Koopman  latent space representation is finite-dimensional a so called Koopman invariant subspace. As a consequence the  Koopman operator and the symmetry generators  admit finite matrix representations. To derive the Koopman representation there exist  elaborate techniques such as e.g.\,\cite{DVK} which have shown significant advantages over a basic variational auto-encoder (VAE). However, we use the latter for simplicity.
Lastly, after having learned a Koopman representation and derived symmetry maps 
the objective is to generate augmented data points; which are then used by the Q-learning algorithm such as CQL at training-time.

\vspace{0.1 cm}
 {\bf The KFC-algorithm }:
\vspace{-0.1 cm}
\begin{itemize}
  \item[] {\bf Step 1:} Train the Koopman forward model to learn the Koopman latent representation.\vspace{0.1 cm}
    \item[] {\bf Step 2:} Compute the symmetry transformations $\sigma_{a}^{\epsilon}$; we use two approaches denoted by {\bf KFC} and {\bf KFC++}.\vspace{0.1 cm}
    \item[] {\bf Step 3:} Integration into Q-learning algorithm; replace input data $ (s_t,s_{t+1},a,r) \mapsto (\sigma_{a}^{\epsilon}(s_t),\sigma_{a}^{\epsilon}(s_{t+1}),a,r) $ .\vspace{0.1 cm}
\end{itemize}

{\bf Step 1: Pre-training of a Koopman forward model }
\beq\label{eq:Fwdmodel}
\mathcal{F}^{c 
 }(s_t,a_t)
    =\begin{cases}
       {\bf  c=0 \;\;\text{VAE}:} \;\;  D\big( E(s_t) \big) = s_t \vspace{0,25 cm}\\ 
       {\bf  c=1 \;\;\text{forward prediction}  :}\vspace{0,1 cm} \\
     D\Big(\, \big(\mathcal{K}_0 + \sum_{i=1}^m \mathcal{K}_i \, a_{t,i} \big) E(s_t) \, \Big) = s_{t+1} 
    \end{cases} \vspace{0,1 cm} \, .
\eeq
Both  $E$ and $D$  are approximated by Multi-Layer-Perceptrons (MLPs) and  the bilinear Koopman-space operator is implemented as fully connected layers (no bias) with weight matrices  $\mathcal{K}_i, \, i=0,\dots,m$, respectively.
 For training details we refer the reader to Appendix \ref{app:implementation}.
 
\vspace{0.2 cm}
{\bf Step 2: Computation of symmetry transformations} 

From theorem \ref{thm:main2} one infers that in order for \Eqref{eq:symoptions} to be a local Lie symmetry group of the approximate ODEs captured by the neural net \Eqref{eq:Fwdmodel} one needs $\sigma_{a_t}$ to commute with the approximate Koopman operator in \Eqref{eq:Fwdmodel}. 
 
For the KFC option in \Eqref{eq:symoptions} the symmetry generator matrix $\sigma_a $ is obtained by solving the equation
$$ \sigma_{a_t} \cdot \mathcal{K}(a_t) - \mathcal{K}(a_t) \cdot\sigma_{a_t}  = 0 \;\; ,$$
which may be accomplished by employing a Sylvester algorithm \citep{sylvester}.
For KFC++  we compute the eigen-vectors of $\mathcal{K}(a_t)$ which then constitute the columns of $U(a_t)$. Thus in particular one infers that 
$$
\big[\;\sigma_{a_t}(\vec\epsilon) \,,\, \mathcal{K}(a_t)\;\big] = 0 \;\;,
$$ 
 The latter, is solved by construction of $\sigma_{a_t}(\vec\epsilon)$ which commutes with the Koopman operator for all values of the random variables.\footnote{ $[ \cdot , \cdot]$ denotes the commutator of two matrices. The Koopman operators eigenvalues and eigen-vectors generically are $\mathbb{C}$-valued.  However, the definition in \Eqref{eq:symoptions} ensures that $\sigma_a(\vec\epsilon)$ is a $\mathbb{R}$-valued matrix.} The symmetry shifts in latent space are given by
\bea\label{eq:symoptions}
&\text{\bf  KFC:}& \;\; {\bf \sigma^\epsilon_{a_t}\hspace{-0.07cm}:}\,s \mapsto \tilde s  =  D \Big(  \big(\mathds{1} + \epsilon  \sigma_{a_t} \big) E(s)  \Big) \\[0.2 cm] \nonumber
&\text{\bf  KFC++:}& \;\; {\bf\sigma^\epsilon_{a_t}\hspace{-0.07cm}:}\,s \mapsto \tilde s = D \Big( \big(\mathds{1} +
\sigma_{a_t}(\vec\epsilon)\big) E(s)  \Big) \, ,\\[0.2 cm] \nonumber
&\sigma_{a_t}(\vec\epsilon)& \hspace{-0.35 cm} =  {\bf \mathscr{R}e} \big(\, U(a_t) \, \text{diag}( \epsilon_1,\dots,  \epsilon_N) \, U^{-1}(a_t)  \big)\;,\;\;
\eea 
where $U(a_t)$ diagonalizes the Koopman operator with eigenvalues $\lambda_i\, ,\; i=1,\dots,N$ i.e.\,the latter can be expressed as 
$$
\mathcal{K}(a_t) =  U(a_t) \, \text{diag}(\lambda_1,\dots, \lambda_N) \, U^{-1}(a_t) \;\; .
$$
Note that we abuse our notation as 
$$E(s) \, \equiv\,  \vec g(s)\,= \,(\,g_1(s),\dots,g_N(s)\,) \;\; , $$
i.e.\,the encoder provides the Koopman space observables. \Eqref{eq:symoptions} holds for both $s_t$ as well as $s_{t+1}$ thus we have deliberately dropped the subscript. The parameters $\epsilon , \, \vec{\epsilon} $ in \Eqref{eq:symoptions} are normally distributed random variables.

Moreover, note that in for KFC++  the symmetry transformations are of the form $\sigma_{a_t}(\vec\epsilon)= \sum_{I=1}\epsilon_I \,\sigma^I_{a_t}$. Lastly,  in the finite-dimensional Koopman invariant sub-spaces matrix multiplication replaces the formal mapping  $\hat \ast$ used in the theorems of section \ref{sec:theory_symmetries}. 

\vspace{0.2 cm}
{\bf Step 3: Integration into a specific Q-learning algorithm (CQL).}

Following  \citep{S4RL}  we leave the policy improvement  \Eqref{eq:policy_improvement3} unchanged but  modify the policy optimisation step \Eqref{eq:policystep} as 
\beq
\hat Q_{i+1} \;\;\;\longleftarrow \; \;\;\;  \tilde\alpha \; \text{Q-regulizer term}\, \;\;\;  {\bf + } \nonumber
\eeq 
\bea
\arg \min_{Q}  \Bigg( \, \underset{ \underset{s_t,s_{t+1}}{a_t,r_t} \sim \mathcal{D}}{\mathbb{E}}\hspace{-0.30cm} &\Bigg[&\hspace{-0.30cm}\Big( r_t + \gamma \, \hat Q_i\Big( \sigma_{a_t}^\epsilon\big(\tilde s_{t+1}|s_{t+1}\big), a_{t+1} \Big) \nonumber\vspace{-0.6 cm}  \\\vspace{-0.8 cm} 
&-& Q\Big( \sigma_{a_t}^\epsilon\big(\tilde s_t|s_t\big), a_t \Big)\Big)^2 \Bigg]\;\Bigg)  \;,
  \label{eq:policystepKFC}
\eea
where  the unmodified Q-regulizer term is given in \Eqref{eq:policystepCQL}.
 
 {\bf Additional remarks:} Symmetry shifts are employed with probability $p_K=0.8$; otherwise a normally distributed state shift is used.
 The KFC and  KFC++ algorithms constitute a simple starting point to extract symmetries. More elaborate studies employing the literature on Koopman spectral analysis are desirable. The reward $r_t$ is not part of the symmetry shift process and will just remain unchanged as an assertion.

\subsection{Limitations}
Note that in practice the assumptions of theorem \ref{thm:main2} hold approximately, e.g.\,$D \circ E \approx id$. The resulting  symmetry transformation shifts both $s_t$ as well as $s_{t+1}$ which evades the necessity of forecasting states. Thus the use of an inaccurate fore-cast model is avoided and the accuracy and generalisation capabilities of the VAE are fully utilized. To limit out-of-distribution generalisation errors the magnitude of the induced shift is controlled by the parameter $\epsilon \ll 1$ such that $|s - \tilde s| = \mathcal{O}(\epsilon)$ . 
KFC is computationally less expensive than KFC++  however it provides less freedom to explore different symmetry directions than the latter.

 Theorems \ref{thm:main} and \ref{thm:main2} imply  that if an operator commutes with the Koopman operator we can find a local symmetry group. However,  global conditions may not be easily inferred, see section \ref{sec:cart} for an example. 
Practically the VAE model parametrized by a neural net is trained on data from one of many behavior policies and is thus learning an approximate dynamics.
The theoretical limitations are twofold, firstly the theorems only hold for dynamical systems with differentiable state-transitions; 
secondly, we employ a Bilinearisation  model for the Koopman operator of the system. 

Many RL environments incorporate dynamics with discontinuous ``contact'' events where the Bilinearisation Ansatz \Eqref{eq:BilinKoop} may not be applicable. 
However, empirically we find that our approach nevertheless is successful for   environments with ``contact'' events. The latter does not affect the performance significantly (see Appendix \ref{app:contact-events}).

\subsection{ Example: Cartpole translation symmetry} \label{sec:cart}
We illustrate the KFC-algorithm for deriving  symmetries from the Koopman latent representation. Moreover, we highlight that this is possible when purposefully employing imprecise  Koopman observables and operators. We refer the reader to the appendix \ref{sec:app_syms} for a more comprehensive discussion. 

The Cartpole environment admits a simple translation symmetry i.e.\,the dynamics are independent of its position $x$.  We choose the Koopman observables to simply be the state vector $g = s =(x,\dot x, \theta, \dot\theta)$; position, angle and their velocities. 
 We empirically train the Koopman forward model  \Eqref{eq:Fwdmodel} on trajectories collected by a set of expert policies. We conclude that the translation symmetry operation emerges when imposing it to be commuting with the Koopman operator. 
 
Note that the Mujoco tasks in D4RL also admit a  position translation symmetry. However,  in the latter case it is common practice to omit the position information from the state vector  to  simplify the policy learning process.  Among other local symmetries our approach can learn those translation symmetries and is thus likely to give additional performance boosts when the position information is not truncated a priori in Mujoco. 
 As most environments do not admit global symmetries  the benefit of our approach is due to utilizing local symmetries.

\begin{table*}[t!]
    \centering
    \caption{We experiment with the full set of the D4RL tasks and report the mean normalized episodic returns over
    5 random seeds using the same protocol as \citet{d4rl}. We compare against 3 competitive baselines including CQL and the two best-performing S4RL-data augmentation strategies. We see that KFC and KFC++ consistently outperforms the baselines. 
    We use the baseline numbers reported in \citet{S4RL}.   \vspace{0.3 cm}}
    \begin{tabular}{l|l | c c c | c c }
        \toprule
         
        \textbf{Domain} & \textbf{Task Name} & \textbf{CQL} & \textbf{S4RL-($\mathcal{N}$)} & \textbf{S4RL-(Adv)} & \textbf{KFC} & \textbf{KFC++}  \\
        \midrule
        \midrule
        \multirow{6}{*}{AntMaze} 
        & antmaze-umaze & 74.0 & 91.3 & 94.1 & \textbf{96.9} & \textbf{99.8} \\
        & antmaze-umaze-diverse & 84.0  & 87.8 & 88.0 & \textbf{91.2}& \textbf{91.1}  \\
        & antmaze-medium-play & 61.2 & 61.9 & 61.6 & 60.0 & \textbf{63.1} \\
        & antmaze-medium-diverse & 53.7 & 78.1 & 82.3 & 87.1& \textbf{90.5} \\
        & antmaze-large-play & 15.8 & 24.4 & \textbf{25.1} & 24.8 & \textbf{25.6} \\
        & antmaze-large-diverse & 14.9 & 27.0 & 26.2 & \textbf{33.1} & \textbf{34.0} \\
        \midrule
        \midrule
        \multirow{12}{*}{Gym} 
        & cheetah-random & 35.4 & \textbf{52.3} & \textbf{53.9} & 48.6 & 49.2  \\
        & cheetah-medium & 44.4 &  48.8 & 48.6 & 55.9 & \textbf{59.1} \\
        & cheetah-medium-replay & 42.0 & 51.4 & 51.7 & \textbf{58.1} & \textbf{58.9}\\
        & cheetah-medium-expert & 62.4 & \textbf{79.0} & 78.1 & \textbf{79.9} & \textbf{79.8}  \\
        & hopper-random & \textbf{10.8} & \textbf{10.8} & \textbf{10.7} & \textbf{10.4} & \textbf{10.7}  \\
        & hopper-medium & 58.0 & 78.9 & 81.3 & 90.6 & \textbf{94.2} \\
        & hopper-medium-replay & 29.5 & 35.4 & 36.8 & \textbf{48.6} & \textbf{49.0} \\
        & hopper-medium-expert & 111.0 & 113.5 & 117.9 & 121.0 & \textbf{125.5}\\
        & walker-random & 7.0 & \textbf{24.9}& \textbf{25.1} & 19.1 & 17.6 \\
        & walker-medium & 79.2 & 93.6 & 93.1 & 102.1 & \textbf{108.0} \\
        & walker-medium-replay & 21.1 & 30.3 & 35.0 & \textbf{48.0} & 46.1\\
        & walker-medium-expert & 98.7 & 112.2 & 107.1 & 114.0 & \textbf{115.3} \\
        \midrule
        \midrule
        \multirow{8}{*}{Adroit} 
        & pen-human & 37.5 &  44.4 & 51.2 & \textbf{61.3} & 60.0 \\
        & pen-cloned & 39.2 & 57.1 & 58.2 & \textbf{71.3} & 68.4\\
        & hammer-human & 4.4 & 5.9 & 6.3 & 7.0 & \textbf{9.4} \\
        & hammer-cloned & 2.1 & 2.7 & 2.9 & 3.0 & \textbf{4.2}\\
        & door-human  & 9.9 & 27.0 & 35.3 & 44.1 & \textbf{46.1} \\
        & door-cloned  & 0.4 & 2.1 & 0.8 & 3.6 & \textbf{5.6} \\
        & relocate-human & \textbf{0.2} & \textbf{0.2} & \textbf{0.2} & \textbf{0.2} & \textbf{0.2} \\
        & relocate-cloned  & \textbf{-0.1} & \textbf{-0.1} & \textbf{-0.1} & \textbf{-0.1}& \textbf{-0.1}\\
        \midrule
        \midrule
        \multirow{2}{*}{Franka} 
        & kitchen-complete & 43.8 & 77.1 & 88.1 & \textbf{94.1} & \textbf{94.9} \\
        & kitchen-partial & 49.8 & 74.8 & 83.6 & 92.3 & \textbf{95.9}  \\
        \bottomrule
    \end{tabular}%
    \label{tab:main_table}
    \vspace{-2pt}
\end{table*}

\section{Empirical evaluation}
\label{emp_results}
In this section, we will first experiment with the popular D4RL benchmark commonly used for offline RL \citep{d4rl}.
The benchmark covers various different tasks such as locomotion tasks with Mujoco Gym \citep{brockman2016openai}, tasks that require hierarchical planning such as antmaze, and other robotics tasks such as kitchen and adroit \citep{adroit}.
Furthermore, similar to S4RL \citep{S4RL}, we perform experiments on 6 different challenging robotics tasks from MetaWorld \citep{metaworld} and RoboSuite \citep{robosuite}.
We compare KFC to the baseline CQL algorithm \citep{CQL}, and two best-performing augmentation variants from S4RL, S4RL-$\mathcal{N}$ and S4RL-adv \citep{S4RL}.
We use the exact same hyperparameters as proposed in the respective papers.
Furthermore, similar to S4RL, we build KFC on top of CQL \citep{CQL} to ensure conservative Q-estimates.

\begin{figure}[t!]
    \centering
    \begin{subcaption}[b]{MetaWorld Environments}
    \includegraphics[width=0.49\textwidth]{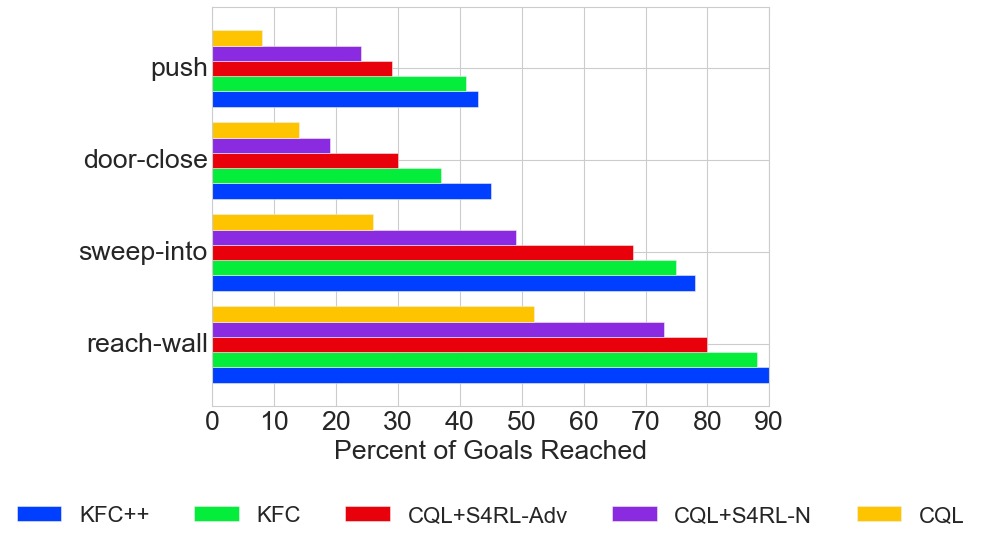}
    \end{subcaption}
    ~
    \begin{subcaption}[b]{RoboSuite Environments}
    \includegraphics[width=0.49\textwidth]{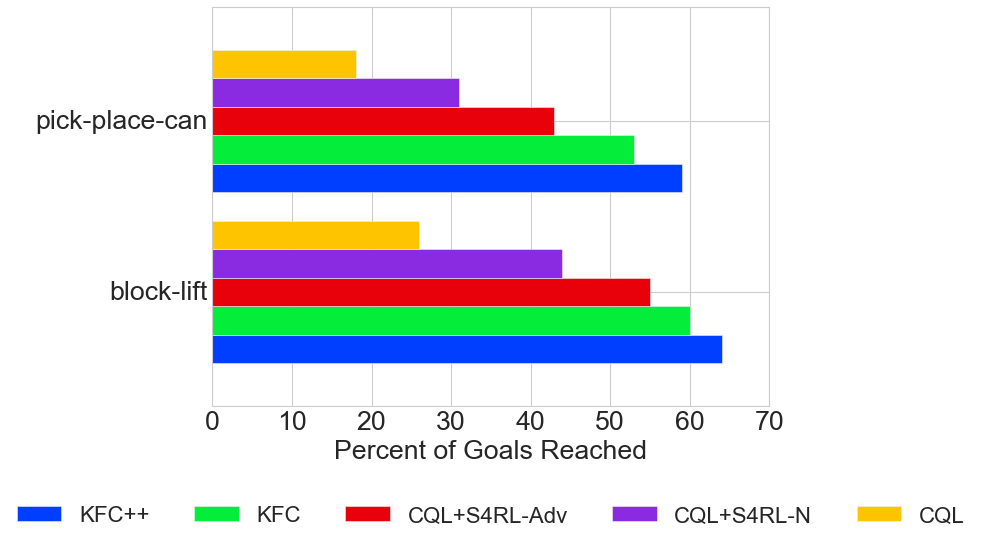}
    \end{subcaption}
    \caption{Results on challenging dexterous robotics environments using data collected using a similar strategy as S4RL \citep{S4RL}. We report the $\%$ of goals that the agent is able to reach during evaluation, where the goal
    is set by the environments.
    \textbf{We see that KFC and KFC++ consistently outperforms both CQL and the two best-performing 
    S4RL variants.}}
    \label{fig:robotics-exps}
\end{figure}

\subsection{D4RL  benchmarks}
\label{sec:Mujoco}

We present results in the benchmark D4RL test suite and report the normalized return in Table \ref{tab:main_table}.
We see that both KFC and KFC++ consistently outperform both the baseline CQL and S4RL across multiple tasks and data distributions. 
Outperforming S4RL-variants on various different types of environments suggests that KFC and KFC++ fundamentally improves the data augmentation strategies discussed in S4RL. 

KFC-variants also improve the performance of learned agents on challenging environments such as antmaze: which requires hierarchical planning, kitchen and adroit tasks: which are sparse reward and have large action spaces.
Similarly, KFC-variants also perform well on difficult data distributions such as ``medium-replay'': which is collected by simply using all the data that the policy encountered while training base SAC policy, and ``-human'' which is collected using human demonstrations on robotic tasks which results in a non-Markovian behaviour policy (more details can be found in the D4RL manuscript \citep{d4rl}).

Furthermore, to our knowledge, the results for KFC++ are state-of-the-art in model-free policy learning from D4RL datasets for most environments and data distributions.
We do note that S4RL outperforms KFC on the ``-random'' split of the data distributions, which is expected as KFC depends on learning a simple dynamics model of the data to use to guide the data augmentation strategy.
Since the ``-random'' split consists of random actions in the environment, our simple  model is unable to learn a useful dynamics model. For ablation studies see appendix \ref{sec:ablation}.




\subsection{Metaworld and Robosuite  benchmarks}

To further test the ability of KFC, we perform additional experiments on challenging robotic tasks.
Following \citep{S4RL}, we perform additional experiments with 4 MetaWorld environments \citep{metaworld}
and 2 RoboSuite environments \citep{robosuite}.
We followed the same method to collect the data as described in Appendix F of S4RL \citep{S4RL}, 
and report the mean percent of goals reached, where the condition of reaching the goal is 
defined by the environment.

We report the results in Figure \ref{fig:robotics-exps}, where we see that by using KFC to guide the 
data augmentation strategy for a base CQL agent, we are able to learn an agent that performs
significantly better.
Furthermore, we see that for more challenging tasks such as ``push'' and ``door-close'' in the 
MetaWorld, KFC++ outperforms the base CQL algorithm and the S4RL agent by a considerable margin.
This further highlights the ability of KFC to guide the data augmentation 
strategy.

\section{Related works}
The use of data augmentation techniques in Q-learning has been discussed recently \citep{NEURIPS2020_e615c82a,pmlr-v119-laskin20a,S4RL}. In particular, our work shares strong parallels with \citep{S4RL}.  Our modification of the policy evaluation step of the CQL algorithm \citep{CQL} is analogous to the one in \citet{S4RL}.  However, the latter randomly augments the data while our augmentation framework is based on symmetry state shifts. 
Regarding the connection to world models \citep{world_models_david} a few comments in order. Here a VAE is used to decode the state information while a recurrent separate neural network predicts future states. Their latent representation is not of Koopman type. Also no symmetries and data-augmentations are derived.

Algebraic symmetries of the state-action space in  Markov Decision Processes  (MDP) originate \citep{homo} an were discussed recently in the context of RL in \citep{vanderPol}.  Their goal is to  preserve the essential algebraic homomorphism symmetry structure of the original MDP  while finding a more compact representation. The symmetry maps considered in our  work are more general and are utilized in a different way. 
Symmetry-based representation learning \citep{higgins}  refers to the study of symmetries of the environment manifested in the latent representation.  
The symmetries in our case are derived form the Koopman operator not the latent representation directly. 
   In \citep{NEURIPS2019_36e729ec} the authors discuss representation learning of symmetries \citep{higgins}  allowing for interactions with the environment. A Forward-VAE model which is similar to our Koopman-Forward VAE model is employed.  We provide theoretical results  on how to derive explicit symmetries of the dynamical systems as well as their utilisation for state-shifts.
\vspace{0.1cm}

 In \citep{sinha2020koopman} symmetries are used to  extend  the Koopman operator from a local to a global description. Neither action-equivariant dynamical control systems nor data augmentation is discussed.
In \citep{doi:10.1063/1.5099091} the imprint of known symmetries on the block-diagonal Koopman space  representation for non-control dynamical systems is discussed. This is close to the spirit of disentanglement \citep{higgins}.  Our results are on control setups and deriving symmetries.  
\vspace{0.1cm}

On another front, the application of Koopman theory in control or reinforcement learning has also been discussed recently. For example, \cite{LHW+20} propose to use compositional Koopman operators using graph neural networks to learn dynamics that can quickly adapt to new environments of unknown physical parameters and produce control signals to achieve a specified goal. \cite{KKB21} discuss the use of Koopman eigenfunction as a transformation of the state into a globally linear space where the classical control techniques is applicable. \cite{IK21} propose a framework for policy learning based on linear quadratic regulator in a space where the Koopman operator is defined. And, \cite{OIL+21} discuss a method of controlling nonlinear systems via the minimization of the Koopman spectrum cost: A cost over the Koopman operator of the controlled dynamics. However, to the best of our knowledge, this paper is the first to discuss Koopman latent space for data augmentation.
  
The study of  Bisimulation metrics of  action-state spaces has lead to performance improvements of Q-learning algorithms \cite{pmlr-v97-gelada19a,zhang2021learning}. We do not alter the metric of the action-state space as well do not take the reward into consideration. It would be interesting to study the KFC-algorithm trained on a  Bisimulation metric space.

\section{Conclusions}
In this work  we have proposed a symmetry-based data augmentation technique derived from a Koopman latent space representation. It enables a principled extension of offline RL  datasets describing dynamical systems.
The approach is based on our theoretical results  on symmetries of dynamical control systems and symmetry shifts of data. Both hold for systems with differentiable state transitions and with a Bilinearisation model for the Koopman operator. However, the empirical results show that the framework is successfully applicable beyond those limitations.

 We empirically evaluated our method on several benchmark offline reinforcement learning tasks D4RL, Metaworld and Robosuite and find that by using  our framework we consistently improve the state-of-the-art of model-free Q-learning algorithms. Our framework is naturally applicable to model-based approaches and thus such an extension of our study would be of great interest.

\subsection*{Acknowledgments}
This work was supported by JSPS KAKENHI (Grant Number JP22H00516), and JST CREST (Grant Number JPMJCR1913).
And, we would like to thank Y.\,Nishimura for technical supports.

\newpage


\bibliography{iclr2022_conference}
\bibliographystyle{icml2022}


\newpage
\appendix
\section{Koopman theory and Symmetries}
\label{app-KTh_and_Sym}
This section provides some additional introductory material on Koopman theory as well as a more detailed presentation of our results on symmetries of dynamical systems.

 In general, a  symmetry group  $\Sigma$ of the state space may be any subgroup of the group of isometrics of the Euclidean space. We  consider Euclidean state spaces  and in particular,  $\Sigma$ invariant compact subsets thereof.
 \begin{definition} [Equivariant Dynamical System]
Consider the dynamical system $\dot{s} = f(s) $ and let $\Sigma$ be a group acting on the state-space $\mathcal{S}$. Then the system is called $\Sigma$-equivariant if $
f (\sigma \cdot s) = \sigma  \cdot f(s) \;, \;\; \text{for} \; s \in \mathcal{S} \;,\;\; \forall \sigma \in \Sigma $.
For a discrete time dynamical system $s_{t+1} = F(s_t)$ one defines equivariance analogously, namely if 
$F (\sigma \cdot s_t) = \sigma  \cdot F(s_t)$ ,  for $\; s_t \in \mathcal{S} \;,\;\; \forall \sigma \in \Sigma $.
\end{definition} 
\vspace{-0.13cm}
 A system of ODEs  may be represented by the set of its solutions which are locally  connected by Lie Groups.
 \begin{definition}[Symmetries of ODEs] A symmetry of a system of ODEs on a  locally-smooth structure is a locally-defined diffeomorphism  that maps the set of all solutions to itself.
\end{definition}
 \vspace{-0.03cm}
 \begin{definition} [Local Lie Group]
\label{def:lieA}
A parametrized set of transformations $\sigma^\epsilon : \mathcal{S} \to \mathcal{S}$  with $s \mapsto \tilde s(s,\epsilon) $ for $\epsilon \in (\epsilon_{low},\epsilon_{high})$ where $\epsilon_{low} < 0 < \epsilon_{high}$ is a one-parameter local Lie group if\footnote{The points (1) and (2) in definition \ref{def:lie} imply the existence of an inverse element  ${\sigma^\epsilon}^{-1} = \sigma^{-\epsilon}$ for $  |\epsilon| \ll 1 $.  A local Lie group satisfies the group axioms for sufficiently small parameters values;  it may not be a group on the entire set. } \vspace{-0.05cm}
\begin{enumerate} \vspace{-0.1cm}
    \item  $\sigma^0$ is the identity map; for $\epsilon=0$ such that $\tilde s(s, 0) = s$.\vspace{-0.1cm}
    \item  $\sigma^{\epsilon_1}\sigma^{\epsilon_2} =\sigma^{\epsilon_2}\sigma^{\epsilon_1} = \sigma^{\epsilon_1 +\epsilon_2}   $ for every $ |\epsilon_1|, |\epsilon_2| \ll 1 $.
   \vspace{-0.1cm} \item $\tilde s(s,\epsilon)$  admits a Taylor series expansion in $\epsilon$; in a neighbourhood of $s$  determined by $\epsilon=0$ as $
     \tilde s(s,\epsilon) = s + \epsilon\,\zeta(s) + \mathcal{O}(\epsilon^2)
    $.
\end{enumerate}
\end{definition}
The Koopman operator of equivariant dynamical system is reviewed in the following .

\begin{lemma}
\label{lem:symA}
The map $ \ast  : \Sigma \times \mathscr{K}(\mathcal{S}) \rightarrow \mathscr{K}(\mathcal{S}) $ given by
$
(\sigma \ast g )(s) \longmapsto  g( \sigma^{-1} \cdot s) $ 
 defines a group action on the Koopman space of observables $\mathscr{K}(\mathcal{S})$. \vspace{-0.cm}
\end{lemma}
\begin{theorem}
\label{thm:SymA}
Let $\mathcal{K}$ be the Koopman operator associated with a $\Sigma$-equivariant system $s_{t+1} = F(s_t)$. Then \vspace{-0.2 cm}
\beq 
[\sigma \ast (\mathcal{K} g)](s) = [\mathcal{K} (\sigma \ast g)](s)  \;\; .
\eeq
\end{theorem}
Theorem~\ref{thm:SymA} states that for a $\Sigma$-equivariant system any symmetry transformation commutes with the Koopman operator. For the proof see \citep{sinha2020koopman}.

Let us next turn to the case relevant for RL, namely control systems. 
In the remainder of this section we focus on dynamical systems given as in \Eqref{eq:nonaffine2} and \Eqref{eq:BilinKoop}.
\begin{definition} [Action Equivariant Dynamical System]
\label{def:act_equi_DSA}
Let $\bar\Sigma$ be a group acting on the state-action-space $\mathcal{S} \times \mathcal{A}$ of a general control system as in \Eqref{eq:nonaffine2} such that it acts as the identity operation on $\mathcal{A}$ i.e.
$
 \sigma\cdot (s_t , a_t) = (\sigma|_{\mathcal{S}} \cdot s_t,a_t) \; ,\;\; \forall \sigma \in \bar\Sigma \;\;.
$
Then the system is called $\bar\Sigma$-action-equivariant if $\forall \sigma \in \bar\Sigma  $
\beq
F (\sigma \cdot(s_t,a_t)) = \sigma|_{\mathcal{S}}  \cdot   F (s_t,a_t) \;, \;\; \text{for} \; (s_t,a_t) \in \mathcal{S}\times \mathcal{A} \; .
\eeq 
\end{definition}

\begin{lemma}
\label{lem:barphiA}
The map $\bar \ast  : \bar \Sigma \times \mathscr{K}(\mathcal{S} \times \mathcal{A} ) \rightarrow \mathscr{K}(\mathcal{S} \times \mathcal{A} ) $ given by
$
(\sigma \bar\ast g )(s,a) \longmapsto  g( \sigma^{-1} \cdot s,a) $ 
 defines a group action on the Koopman space of observables $\mathscr{K}(\mathcal{S} \times \mathcal{A} )$. 
\end{lemma}
\begin{theorem}
\label{thm:Sym2A}
A Koopman operator $\mathcal{K}$  of a $\bar\Sigma$-action-equivariant system $s_{t+1} = F(s_t,a_t)$ satisfies  
\beq
[\sigma \bar\ast (\mathcal{K} g)](s_t,a_t) = [\mathcal{K} (\sigma \bar\ast g)](s_t,a_t) \;\; .
\eeq 
\end{theorem}\vspace{-0.1 cm}
In particular, it is easy to see that the biliniarisation in \Eqref{eq:BilinKoop} is  $\Sigma$-action-equivariant if
$
f_i(\sigma|_{\mathcal{S}} \cdot s) = \sigma|_{\mathcal{S}} \cdot f_i(s)\;,\;\; \forall i=0,\dots,m
$.
Let us thus turn our focus on the  relevant case of a control system    $s_{t+1} = \tilde F (s_t,a_t)$ which admits a Koopman operator description as 
\beq\label{eq:KAA}
g(s_{t+1}) = \mathcal{K}(a_t)g(s_{t}) \;,\;\; \text{for}\; a_t\in \mathcal{A} \,, \;\forall g \in \mathscr{K}(\mathcal{S})\;\;,
\eeq
where $\{ \mathcal{K}(a) \}_{a \in \mathcal{A}}$ is a family of operators with analytical dependence on $a \in \mathcal{A}$.
Note that the bilinearisation in \Eqref{eq:BilinKoop} is a special case of \Eqref{eq:KA}.
Furthermore, let $\{U(a)\}_{a \in \mathcal{A}}$ be a family of invertible operators s.t. $U(a): \mathscr{K}(\mathcal{S}) \to \mathscr{F}(\mathcal{S} \times \mathcal{A})$ is a mapping to the (Banach) space of eigenfunctions $ \mathscr{F}(\mathcal{S} \times \mathcal{A})$ with eigenfunctions  $\varphi(s,a) := U(a)g(s)$ which obeys  $ U(a) \mathcal{K} (a) U(a)^ {-1} = \Lambda(a)$, with  $\Lambda(a) \varphi(s,a)  = \lambda_\varphi(a)\varphi(s,a) $ and where $\lambda_\varphi(a): \mathcal{A} \to \mathbb{R}$. The existence of such operators on an infinite dimensional space requires the Koopman operator  in \Eqref{eq:KA}  to be self-adjoint or a finite-dimensional (approximate) matrix representation to  be diagonalizable.\footnote{See Appendix~\ref{app-KTh_and_Sym} for the details. We found that, in practice the Koopman operator leaned by the neural nets is diagonalizable almost everywhere. }
\vspace{0.1 cm}
\begin{lemma} 
\label{lem:hatphiA}
The map $\hat\phi : \bar \Sigma \times \mathscr{K}(\mathcal{S}) \times \{U(a)\}_{a \in \mathcal{A}} \rightarrow \mathscr{K}(\mathcal{S} ) $ given by
\beq
(\sigma_a \hat\ast g )(s) \longmapsto  \Big( U^{-1}(a) \big( \sigma\bar\ast ( U(a) g) \big) \Big) (s)= g(  \sigma^{-1} \cdot s) \;\;,
\eeq
 defines a group action on the Koopman space of observables $\mathscr{K}(\mathcal{S})$.  Where $\bar\ast$ is defined analog to Lemma \ref{lem:barphiA} but acting on $\mathscr{F}(\mathcal{S} \times \mathcal{A})$ instead of $\mathscr{K}(\mathcal{S} \times \mathcal{A})$ by $
(\sigma \bar\ast \varphi )(s,a) \longmapsto  \varphi ( \sigma^{-1} \cdot s,a)$. We refer to a control system  in \Eqref{eq:KAA} admitting a $\hat\phi$-symmetry as $\hat\Sigma$-action-equivariant.
\end{lemma}

\begin{theorem}
\label{thm:mainA}
Let $s_{t+1} = \tilde F (s_t,a_t)$ be a $\hat\Sigma$-action-equivariant  control system with a symmetry action as in Lemma \ref{lem:hatphiA} which furthermore admits a Koopman operator representation as
\beq\label{eq:goodsysA}\vspace{-0.05cm}
g(s_{t+1}) = \mathcal{K}(a_t)g(s_{t}) \;,\;\; \text{for}\; a_t\in \mathcal{A} \,, \;\forall g \in \mathscr{K}(\mathcal{S}) \;\;.
\eeq
Then \vspace{-0.05cm}
\beq
\label{eq:mainrelA}
\big[\sigma_{a_t} \hat\ast \big(\mathcal{K}(a_t) g\big) \big](s_t) =\big[\mathcal{K}(a_t) \, \big(\sigma_{a_t}  \hat\ast g \big)\big](s_t) \;\;.
\eeq
Moreover, a control system obeying equations~\ref{eq:goodsysA} and \ref{eq:mainrelA} is $\hat\Sigma$-action-equivariant locally if $g^{-1}$ exists for a neighborhood of $s_t$, i.e.\,then $\sigma \cdot \tilde F(s_t,a_t) =  \tilde F( \sigma \cdot (s_t,a_t))$.
\end{theorem}
 To allow an easier transition to the empirical section  we introduce  the notation  $E: \mathcal{S} \to \mathscr{K}(S)$ and $D: \mathscr{K}(S) \to \mathcal{S}$ denote the $\mathcal{C}^1$-differentiable encoder and decoder to and from the finite-dimensional Koopman space approximation, respectively, i.e.\,$E \circ D = D \circ E = id$. 
 Data-points may be shifted by symmetry transformations of solutions of ODEs as discussed next.

\begin{theorem}
\label{thm:main2A}
Let $s_{t+1} = \tilde F (s_t,a_t)$ be a control system as in \Eqref{eq:goodsysA} and $\sigma_{a_t}$ an operator obeying \Eqref{eq:mainrelA}. 
Then $\sigma_{a_t}^\epsilon : \big(s_t,s_{t+1},a_t\big)  \longmapsto  \big(\tilde s_t, \tilde s_{t+1},a_t\big) $ with 
\bea\label{eq:symshiftA}
\tilde s_t &=& D\Big(\,\big(\mathds{1} \;+\; \epsilon \, \sigma_{a_t}\big)\hat\ast E(s_t) \Big) \; ,\\ 
\nonumber  \tilde s_{t+1} &=& D\Big(\,\big(\mathds{1} \;+\; \epsilon \,\sigma_{a_t} \big) \hat\ast E(s_{t+1}) \Big) \;\; \,
\eea
is a one-parameter local Lie symmetry group of ODEs.
In other words one can use a symmetry transformation to shift both $s_t$ as well as $s_{t+1}$ such that  $\tilde s_{t+1} = \tilde F (\tilde s_t,a_t)$.
\end{theorem}
No assumptions on the Koopman operator are imposed. Moreover, note that the equivalent theorem holds when $\epsilon \,\sigma_{a_t} \to \sum_{I=1}\epsilon_I \,\sigma^I_{a_t}$ to be a local N-parameter Lie group.

{\bf Additional Remarks.} Let us re-evaluate some information given on the operators used previously.  Let $\{U(a)\}_{a \in \mathcal{A}}$ be a family of invertible operators s.t. $U(a): \mathscr{K}(\mathcal{S}) \to \mathscr{F}(\mathcal{S} \times \mathcal{A})$ is a mapping to the (Banach) space of eigenfunctions  $\varphi(s,a) := U(a)g(s) \in \mathscr{F}(\mathcal{S} \times \mathcal{A})$. Which moreover obeys  
\bea
U(a) \mathcal{K} (a) U(a)^ {-1} &=& \Lambda(a), \;\; \text{with} \\\nonumber \Lambda(a) \varphi(s,a)  &=& \lambda_\varphi(a)\varphi(s,a) \;\; , 
\eea
and where $\lambda_\varphi(a): \mathcal{A} \to \mathbb{R}$. The existence of such operators puts further restriction on the Koopman operator in \Eqref{eq:KA}.  However, as our algorithm employs the finite-dimensional approximation of the Koopman operator i.e.\,its  matrix representation this amounts simple for $\mathcal{K}(a)$ to be diagonalizable and $U(a)$ is the matrix containing its eigen-vectors as columns.
To evolve a better understanding on the required criteria for the infinite-dimensional case we employ  an alternative formulation of the  so called spectral theorem below.
\begin{theorem}[Spectral Theorem]
 Let $\mathcal{K}$ be a bounded self-adjoint operator on a Hilbert space $\mathcal{H}$. Then there is a measure space $(\mathcal{S}, \Sigma, \mu)$ and a real-valued essentially bounded measurable function $\lambda$ on $\mathcal{S}$ and a unitary operator $U: \mathcal{H} \to L^{2}_\mu(\mathcal{S})$ i.e.\,$U^\ast  U = U \, U^\ast = id$ such that
 \beq
 U \, \Lambda  \,  U^\ast = \mathcal{K} \;,\;\;\text{with}\;\; \big [\Lambda \varphi \big](s) = \lambda(s)\varphi(s) \;\;.
 \eeq
\end{theorem}
In other words  every bounded self-adjoint operator is unitarily equivalent to a multiplication operator.
In contrast to the finite-dimensional case we need to slightly alter our criteria to $ U(a) \mathcal{K} (a) U(a)^ {-1} = \Lambda(a)$, with  $\Lambda(a) \varphi(s,a)  = \lambda_\varphi(a,s)\varphi(s,a) $ and where $\lambda_\varphi(s,a): \mathcal{S} \times \mathcal{A} \to \mathbb{R}$.
Concludingly, a sufficient condition for our criteria  to hold in terms of operators on Hilbert spaces is that the Koopman operator $\mathcal{K}(a)$ is self-adjoint i.e.\,that
\beq
\mathcal{K}(a) = \mathcal{K}(a)^ {\ast} \;\; .
\eeq
\section{Proofs}
\label{app:proofs}
In this section we provide the proofs of the theoretical results of Section \ref{KFC_algo}.
\paragraph{Proof of Lemma \ref{lem:symA} and Theorem \ref{thm:SymA}}
The proofs of the lemma as well as the theorem can be found in  \citep{sinha2020koopman}.
\paragraph{Proof of Lemma \ref{lem:barphi}:}
We aim to show that map $\bar \ast  : \bar \Sigma \times \mathscr{K}(\mathcal{S} \times \mathcal{A} ) \rightarrow \mathscr{K}(\mathcal{S} \times \mathcal{A} ) $ given by
$
(\sigma \bar\ast g )(s,a) \longmapsto  g( \sigma^{-1} \cdot s,a) $ 
 defines a group action on the Koopman space of observables $\mathscr{K}(\mathcal{S} \times \mathcal{A} )$, where  $\bar \Sigma$ defines the symmetry group of definition \ref{def:act_equi_DS}.
 Firstly, let $g \in \mathscr{K}(\mathcal{S} \times \mathcal{A} )$ and $\sigma_0 \in \bar \Sigma$  be the identity element, then we see that it provides existence of an identity element of $\bar \ast $ by
 $$
 (\sigma_0 \bar\ast g) (s,a) = g\big(\sigma_0^{-1} \cdot s, a \big) = g(s,a) \;\;.
 $$
 Secondly, let $\sigma_1, \sigma_2 \in \bar\Sigma$ and $\odot $ denoting the group operation i.e.\,$\sigma_1 \odot\sigma_2 = \sigma_3\in \bar\Sigma$. Then 
 \bea
 \Big( \sigma_2 \bar \ast \big(\sigma_1 \bar \ast g \big) \Big) (s,a) &=& \sigma_2 \bar \ast \Big( g \big(\sigma_1^ {-1} \cdot s,a\big)\Big)  \nonumber
 \\[0,2 cm] \nonumber
  &= &  g \big(\sigma_2^ {-1}\cdot ( \sigma_1^ {-1} \cdot s ),a\big)  \\[0,2 cm] \nonumber
  &=&
  g \big( (\sigma_2^ {-1}\odot \sigma_1^ {-1}) \cdot s ),a\big)    \nonumber \\[0,2 cm] \nonumber
  & \stackrel{(I)}{ =}& g \big( (\sigma_1 \odot \sigma_2)^ {-1} \cdot s,a\big) 
  \\[0,2 cm]
 &=&\Big(  \big( \sigma_1 \odot \sigma_2 \big)\bar \ast g  \Big) (s,a) \nonumber
 \\[0,2 cm]
 &=& \Big(  \sigma_3 \bar \ast g  \Big) (s,a) \;\; ,
 \eea
 where in $(I)$ we have used the invertibility of the group property of $\bar \Sigma$. Lastly it follows analogously that for $\sigma , \sigma^{-1} \in \bar \Sigma$ that 
 \bea
 \Big( \sigma \bar \ast \big(\sigma^{-1} \bar \ast g \big) \Big) (s,a)   &=& \Big(  \big( \sigma^{-1} \odot \sigma \big)\bar \ast g  \Big) (s,a) \\[0,2 cm]\nonumber
 &=& \Big(  \sigma_0 \bar \ast g  \Big) (s,a) =  g(s,a) \;\; . \nonumber
 \eea
Thus the existence of an inverse is established which concludes to show the group property of $\bar \ast $.

\paragraph{Proof of Theorem \ref{thm:Sym2}:}
We aim to show that with $\mathcal{K}$  being the Koopman operator associated with a $\bar\Sigma$-action-equivariant system $s_{t+1} = F(s_t,a_t)$. Then
$$
\Big[\, \sigma \bar\ast \big(\mathcal{K} g \big)\,\Big](s,a) = \Big[\,\mathcal{K} \big(\sigma \bar\ast g \big)\Big](s,a) \;\; .
$$
First of all note that by the definition of the Koopman operator of non-affine control systems it obeys $$[ \mathcal{K} g](s_t,a_t) = g(F(s_t,a_t),a_{t+1})\;\;.$$ Using the latter one thus infers that
\bea
\Big[ \sigma \bar\ast \big(\mathcal{K} g\big)\Big](s_t,a_t)  &=&  \sigma \bar\ast  g(F(s_t,a_t),a_{t+1}) 
\\[0,2 cm]\nonumber
&=&   g(\sigma^ {-1}  \cdot F(s_t,a_t),a_{t+1})
\\[0,2 cm]\nonumber
&\stackrel{(I)}{=}& g(F(\sigma^ {-1} \cdot s_t,a_t),a_{t+1}) \,,\nonumber
\eea
where in (I) we have used that it is a $\bar\Sigma$-action-equivariant system. Moreover, one finds that
\bea
 g(F(\sigma^{-1} \cdot s_t,a_t),a_{t+1})  &=&  \mathcal{K} g(\sigma^{-1} \cdot s_t,a_t) \\[0,2 cm]\nonumber
&=&   \Big[\,\mathcal{K} \big(\sigma \bar\ast g\big) \,\Big](s_t,a_t)\;\;, \nonumber
\eea
which concludes the proof.

\paragraph{Proof of Lemma \ref{lem:hatphiA}:}
We aim to show that the map $\hat\phi : \bar \Sigma \times \mathscr{K}(\mathcal{S}) \times \{U(a)\}_{a \in \mathcal{A}} \rightarrow \mathscr{K}(\mathcal{S} ) $ given by
\beq\nonumber
\big(\sigma_a \hat\ast g \big)(s) \longmapsto  \Big( U^{-1}(a) \big( \sigma\bar\ast ( U(a) g) \big) \Big) (s)= g\big(  \sigma^{-1} \cdot s\big) \;\;,
\eeq
 defines a group action on the Koopman space of observables $\mathscr{K}(\mathcal{S})$.  Where $\bar\ast$ is defined analog to Lemma \ref{lem:barphi} but acting on $\mathscr{F}(\mathcal{S} \times \mathcal{A})$ instead of $\mathscr{K}(\mathcal{S} \times \mathcal{A})$ by $
(\sigma \bar\ast \varphi )(s,a) \longmapsto  \varphi ( \sigma^{-1} \cdot s,a)$. 
First of all note that
\bea\label{eq:phihat0}
 \Big( U^{-1}(a) \big( \bar{\sigma}\ast ( U(a) g) \big) \Big) (s)&=&  U^{-1}(a)  \big( \bar{\sigma}\ast  \varphi(s,a) \big)\nonumber \\[0,2 cm]\nonumber
&=& U^{-1}(a) \varphi\big(\sigma^{-1} \cdot s,a\big)  
 \\[0,2 cm]
&=& g\big(  \sigma^{-1} \cdot s\big)
\eea
We proceed analogously as in the proof of Lemma \ref{lem:barphi} above.
 Firstly, let $g \in \mathscr{K}(\mathcal{S})$ and  $\sigma_0 \in \bar \Sigma$ be the identity element then on infers from \Eqref{eq:phihat0} that
it provides existence of an identity element of $\hat\phi$ by
 $$
 (\sigma_{a,0} \hat\ast g) (s) = g\big(\sigma_0^{-1} \cdot s \big) = g(s) \;\;,
 $$
 where we have used the notation $\text{for} \;\; i=0,\dots $
 $$
 (\sigma_{a,i} \hat\ast g) (s) = \Big( U^{-1}(a) \big( \sigma_i\bar\ast ( U(a) g) \big) \Big) (s) \;, \;\;.
 $$
 Secondly, let $\sigma_1, \sigma_2 \in \bar\Sigma$ and $\odot $ denoting the group operation i.e.\,$\sigma_1 \odot\sigma_2 = \sigma_3\in \bar\Sigma$. Then 
 \bea
 \Big( \sigma_{a,2} \hat \ast \big(\sigma_{a,1} \hat \ast g \big) \Big) (s)  &=& \sigma_{a,2} \hat \ast \Big( g \big(\sigma_1^ {-1} \cdot s\big)\Big)  
 \\[0,2cm]\nonumber
 &=& g \big(\sigma_2^ {-1}\cdot ( \sigma_1^ {-1} \cdot s )\big)  
 \\[0,2cm]\nonumber
 &=& g \big( (\sigma_2^ {-1}\odot \sigma_1^ {-1}) \cdot s \big)    \nonumber 
 \\[0,2 cm]\nonumber
 &\stackrel{(I)}{ =}& g \big( (\sigma_1 \odot \sigma_2)^ {-1} \cdot s\big)  
 \\[0,2cm]\nonumber
 &=&
 g \big( \sigma_3^ {-1} \cdot s\big) 
  \\[0,2cm]\nonumber
 &=&\Big(  \sigma_{a,3} \bar \ast g  \Big) (s) \;\; ,
 \nonumber
 \eea
 where in $(I)$ we have used the invertibility of the group property of $ \Sigma$.
 Lastly, it follows analogously that for $\sigma , \sigma^{-1} \in \bar \Sigma$ that 
 \bea
 \Big( \sigma \hat \ast \big(\sigma^{-1} \hat \ast g \big) \Big) (s)   &=&\Big(  \big( \sigma^{-1} \odot \sigma \big)\hat \ast g  \Big) (s) \\[0,2cm]\nonumber
 &=& \Big(  \sigma_0 \hat \ast g  \Big) (s) =  g(s) \;\; . \nonumber
 \eea
Thus the existence of an inverse is established which concludes to show the group property of $\hat \phi$.

\paragraph{Proof of Theorem \ref{thm:main}:}
We aim to show twofold.
\paragraph{\underline{$\Longrightarrow$}:} Firstly, that with $s_{t+1} = \tilde F (s_t,a_t)$ be a $\hat\Sigma$-action-equivariant  control system with a symmetry action as in Lemma \ref{lem:hatphiA} which furthermore admits a Koopman operator representation as
\beq
g(s_{t+1}) = \mathcal{K}(a_t)g(s_{t}) \;,\;\; \text{for}\; a_t\in \mathcal{A} \,, \;\forall g \in \mathscr{K}(\mathcal{S}) \;\;.\nonumber
\eeq
Then
\beq
\Big[\sigma_{a_t} \hat\ast \big(\mathcal{K}(a_t) g\big) \Big](s_t) =\Big[\mathcal{K}(a_t) \, \big(\sigma_{a_t} \hat\ast g \big)\Big](s_t) \;\;. \nonumber
\eeq
\paragraph{\underline{$\Longleftarrow$}:}
Secondly, the converse. Namely, if a control system $s_{t+1} = \tilde F (s_t,a_t)$ obeys Eqs.~(\ref{eq:goodsys}) and (\ref{eq:mainrel}), then
it is $\hat\Sigma$-action-equivariant, i.e.\,$\sigma \cdot \tilde F(s_t,a_t) =  \tilde F( \sigma \cdot (s_t,a_t)) $. For notational simplicity we drop the subscripts referring to time in the remainder of this proof i.e.\, $s_t \to s$  and $a_t \to a$.

Let us start with the first implication i.e.\,$\Longrightarrow$.

First of all note that by the definition of the Koopman operator one has  $$\big[ \mathcal{K}(a) g \big](s) = g\big(\tilde F(s,a)\big)\;\;.$$ Using the latter one infers that
\bea
\Big[ \sigma_a \hat\ast \big(\mathcal{K}(a) g\big)\Big](s) &=& \sigma_a \hat\ast  g(\tilde F(s,a))  \nonumber \;\; ,\\[0,2 cm]
&=&  U^{-1}(a) \Big( \sigma\bar\ast \big( U(a) g\big(\tilde F(s,a) \big) \big) \Big)  \nonumber \;\; ,\\[0,2 cm]
&=&  U^{-1}(a) \Big( \sigma\bar\ast  \varphi\big(\tilde F(s,a),a \big) \Big)  \nonumber \;\; ,\\[0,2 cm]
&=&  U^{-1}(a) \Big(  \varphi\big(\sigma^{-1} \cdot \tilde F(s,a),a \big) \big) \Big)  \nonumber \;\; ,\\[0,2 cm]
&=&  U^{-1}(a) \Big(  \varphi\big( \tilde F\big(\sigma^{-1} \cdot s,a\big),a \big) \big) \Big)  \nonumber \;\; ,\\[0,2 cm]
&=&  g \big( \tilde F\big(\sigma^{-1}s,a\big)\big)  \nonumber \;\; .
\eea
Moreover, one derives that
\bea
 g \big( \tilde F\big(\sigma^{-1}s,a\big)\big)  &=&  \mathcal{K}(a)g\big( \sigma^{-1} s\big) \nonumber \;\; ,\\[0,2 cm]
&=&   \mathcal{K}(a) \, \underbrace{U^{-1}(a)\, U(a)}_{=id} g  \big( \sigma^{-1} s\big) \nonumber \;\; ,\\[0,2 cm]
&=&   \mathcal{K}(a) \, U^{-1}(a)\, \varphi \big( \sigma^{-1} s,a\big) \nonumber \;\; ,\\[0,2 cm]
&=&   \mathcal{K}(a) \Big( \, U^{-1}(a)\,\big(\sigma \bar \ast \varphi \big( s,a\big) \big) \Big) \nonumber \;\; ,\\[0,2 cm]
&=&   \mathcal{K}(a) \Big( \, U^{-1}(a)\,\big( \sigma \bar \ast \big( U(a) g( s) \big) \big) \Big) \nonumber \;\; ,\\[0,2 cm]
&=&\Big[\mathcal{K}(a) \, \big(\sigma_a \hat\ast g \big)\Big](s) \nonumber \;\; ,
\eea
which concludes the proof of the first part of the theorem.
Let us next show the converse implication i.e.\,$\Longleftarrow$. For this case it is practical to use the discrete time-system notation explicitly. Let $\sigma \in \Sigma$ be a symmetry of the state-space and be $\tilde s_t = \sigma \cdot s_t$  and $\tilde s_{t+1} = \sigma \cdot s_{t+1}$ the $\sigma$-shifted states. Then
\bea
 g \big(s_{t+1}\big)  &=&   g \big(\tilde F(s_t,a_t)\big) \nonumber \;\; ,\\[0,2 cm]
  &=&   \mathcal{K}(a_t) \, g (s_t) \nonumber \;\; ,\\[0,2 cm]
  &=&   \mathcal{K}(a_t)\,  g \big(\sigma^{-1}\cdot \tilde s_t\big) \nonumber \;\; ,\\[0,2 cm]
    &=&   \mathcal{K}(a_t) \Big( \sigma_a \hat\ast g \big( \tilde s_t\big) \Big) \nonumber \;\; ,\\[0,2 cm]
     &=&  \Big[ \mathcal{K}(a_t) \big( \sigma_a \hat\ast g \big)\Big] ( \tilde s_t )  \nonumber \;\; ,\\[0,2 cm]
  &\stackrel{(I)}{=}&  \Big[\sigma_a \hat\ast \big(\mathcal{K}(a_t) g\big) \Big](\tilde s_t) \nonumber \;\; ,
\eea
where in (I) we have used that the symmetry operator commutes with the Koopman operator.
Thus in particular
\bea
 \sigma_a^{-1} \hat\ast g \big(s_{t+1}\big)  &= & \Big[ \sigma_a^{-1} \hat\ast  \big(\sigma_a \hat\ast \big(\mathcal{K}(a_t) g\big) \big)\Big](\tilde s_t)\nonumber 
 \\[0,2 cm]
  &\stackrel{Lemma \; \ref{lem:hatphiA}}{=}  & \Big[ \mathcal{K}(a_t) g \Big](\tilde s_t) \;\; ,
 \eea
 Moreover, one finds that
\bea
\sigma_a^{-1}\hat \ast g \big(s_{t+1}\big) &=&  g \Big(\big(\sigma^{-1}\big)^{-1}\cdot s_{t+1}\Big) \\[0,2 cm]
&=&   g \big(\sigma\cdot s_{t+1}\big) =  g \big(\tilde s_{t+1}\big) \;\;,
\nonumber
\eea
from which one concludes that
\beq
g (\tilde s_{t+1}) =  \big[ \mathcal{K}(a_t) g \big](\tilde s_t) = g\big(\tilde F(\tilde s_t,a_t)\big) \;\; .
\nonumber
\eeq
Finally, we  use  the invertibility of the Koopman space observables i.e.\,$g^{-1}(g(s)) = s$  to infer
\beq
\tilde s_{t+1} =  \tilde F(\tilde s_t,a_t) \nonumber \;\;.
\eeq
However, in general $s$ the Koopman space observables are not invertible globally. The "inverse function theorem"  guarantees the existence of a local inverse if $g(s)$ is $\mathcal{C}^1$ differentiable for maps between manifolds of equal dimensions; see any classical reference/text-book on differential calculus. However, we assume inevitability locally.
\bea
\tilde s_{t+1} = \sigma \cdot s_{t+1} \;\; &\Rightarrow& \;\;\tilde F(\tilde s_t,a_t) =  \sigma \cdot \tilde F(s_t,a_t) \;\; 
\\[0,2 cm]
     &\Rightarrow& \tilde F(\sigma \cdot s_t,a_t) =  \sigma \cdot \tilde F(s_t,a_t) \;\;,\nonumber
\eea
which at last concludes our proof of the second part of the theorem.
\paragraph{Extension of Theorem \ref{thm:main2}:} Moreover, one may account for practical limitations i.e.\,an error by the assumption  $[\sigma_{a_t}^\epsilon, \mathcal{K}(a_t) ]= \epsilon_a \, \mathds{1}$. One then finds that $\tilde s_{t+1} = \tilde F (\tilde s_t,a_t) + \mathcal{O}(\epsilon_a)$. Thus the error becomes suppressed when $\epsilon_a \ll \epsilon$. The error may be due to practical limitations of capturing the true dynamics as well as the symmetry map.
\paragraph{Proof of Theorem \ref{thm:main2}:} We will show here theorem \ref{thm:main2} as well as the extension mentioned in the paragraph prior to this proof.
Let $s_{t+1} = \tilde F (s_t,a_t)$ be a $\hat\Sigma$-action-equivariant  control system as in Theorem \ref{thm:main}. For symmetry maps
\bea\label{proofsym}
\tilde s_t &=& D\Big(\,\big(\mathds{1} \;+\; \epsilon \, \sigma_{a_t} \big)\hat\ast E(s_t) \Big) \; , \;\; \\[0,2 cm] \nonumber 
  \tilde s_{t+1}  &=& D\Big(\,\big(\mathds{1} \;+\; \epsilon \,\sigma_{a_t} \big) \hat\ast E(s_{t+1}) \Big) \;\; .
\eea
we aim to show that one can use a symmetry transformation to shift both $s_t$ as well as $s_{t+1}$
\beq
\sigma_{a_t}^\epsilon : \;\; \big(s_t,s_{t+1},a_t\big) \;\;  \longmapsto  \;\; \big(\tilde s_t, \tilde s_{t+1},a_t\big) \eeq 
such that 
\beq
\tilde s_{t+1} = \tilde F (\tilde s_t,a_t) \; .
\eeq 
From \Eqref{proofsym} and the definition of the Koopman operator one infers that\footnote{For notational simplicity we study the general case  $(\mathds{1} \;+\; \epsilon \, \sigma_{a_t} )  \to\sigma_{a_t}$. }
\beq\label{st-proof}
\tilde s_{t+1}=  D\Big(\ \sigma_{a_t}\hat\ast E( s_{t+1}) \Big) =  D\Big(\ \sigma_{a_t}\hat\ast \mathcal{K}(a_t) E( s_{t}) \Big)  \;\;.
\eeq
By using that $[\sigma_{a_t},\mathcal{K}(a_t)] = 0$ and \Eqref{proofsym} one finds that
\bea
\tilde s_{t+1}&=& D\Big(\ \mathcal{K}(a_t)\sigma_{a_t}\hat\ast E( s_{t}) \Big) \\[0,2 cm] \nonumber 
  &=&  D\Big(\ \mathcal{K}(a_t) \underbrace{E( D(}_{=\mathds{1}} \sigma_{a_t}\hat\ast E( s_{t}))) \Big)
  \\[0,2 cm] \nonumber 
  &=& D\Big(\ \mathcal{K}(a_t) E(\tilde s_{t}) \Big)   \;\;,
\eea
which concludes the proof of the first part of the theorem with $\sigma_{a_t} \to (\mathds{1} \;+\; \epsilon \, \sigma_{a_t}  )$. 
\newline
{\bf Local diffeomorphism. }
Per definition the maps $D,E$ are differentiable and invertible which implies that they provide a local diffeomorphism from and to Koopman-space. 
Also the linear map i.e.\,a matrix multiplication by $(\mathds{1} \;+\; \epsilon \, \sigma_{a_t}  )$ is a diffeomorphism. Thus the symmetry map  $\sigma_{a_t}^\epsilon$ i.e.\,\Eqref{eq:symshift} constitutes a local diffeomorphism. 
The above proof culminating in \Eqref{proofsym}  implies that we have a local diffeomorphism mapping solutions of the ODEs to solutions, thus a symmetry of the system of ODEs.
\newline
{\bf Limitations. }
However $D,E$ to be invertible is a strong assumption as it implies that the Koopman space approximation admit the same dimension as the state space. Note that however in practice as we only require  $D \circ E \approx id$ one may choose other Koopman space dimensions.
\newline
{\bf Local Lie group. }
What is left to show is that the symmetry of the system of ODEs locally is a Lie group. In definition \ref{def:lie} we need to show points (1)-(3). 
\begin{enumerate} \vspace{-0.1cm}
    \item For $\epsilon=0$ one finds that $\sigma^0_{a_t}$ is the identity map i.e for such that 
    \beq
    \tilde s_t(s_t, 0) = D\big( (\mathds{1} \;+\; 0 \cdot \, \sigma_{a_t})E(s_t)\big) = D( E(s_t) ) = s_t
    \eeq
    \item  
    \bea \sigma^{\epsilon_1}\sigma^{\epsilon_2} \hspace{-0.2cm} &=& \hspace{-0.2cm}  D\big( (\mathds{1} + \epsilon_1 \sigma_{a_t} )(\mathds{1} + \epsilon_2 \sigma_{a_t} )E(s_t)\big)  \\ \nonumber
     &=& \hspace{-0.2cm} D\big( (\mathds{1} + (\epsilon_1  + \epsilon_2) \sigma_{a_t}  + \mathcal{O}(\epsilon_1 \cdot \epsilon_2))E(s_t)\big) \\ \nonumber
       &=& \hspace{-0.2cm} \sigma^{\epsilon_1 +\epsilon_2}  
    \eea
    for every $ |\epsilon_1|, |\epsilon_2| \ll 1 $.
   \item $\tilde s_t(s_t,\epsilon)$  admits a Taylor series expansion in $\epsilon$, i.e.\,in a neighbourhood of $s$  determined by $\epsilon=0$. We may Taylor expand D around the point $E(s_t) \equiv g$ as
   \beq
   \tilde s_t(s_t,\epsilon) = D(E(s_t)) + \epsilon \sum_{I=1}^N (\sigma_{a_t} E(s_t))_I \frac{\partial D}{\partial g_I} + \mathcal{O}(\epsilon^2)
   \eeq
   thus
   \beq
     \tilde s_t(s_t,\epsilon) = s_t + \epsilon\zeta(s_t) + \mathcal{O}(\epsilon^2)
    \eeq
    for $\zeta(s_t) = \sum_{I=1}^N (\sigma_{a_t} E(s_t))_I \tfrac{\partial D}{\partial g_I}$.
    \item The existence of an inverse element $(\sigma_{a_t}^\epsilon)^{-1} =\sigma_{a_t}^{-\epsilon}$ follows from (1) and (2).
\end{enumerate} 

{\bf Numerical errors. } Let us next turn to the second part of the theorem incorporating for practical numerical errors. We aim to show that under the assumption $[\sigma_{a_t}^\epsilon , \mathcal{K}(a_t)]= \epsilon_a \, \mathds{1}$ one finds  $\tilde s_{t+1} = \tilde F (\tilde s_t,a_t) + \mathcal{O}( \epsilon_a) $.
Note that $\sigma_{a_t} \to (\mathds{1} \;+\; \epsilon \, \sigma_{a_t} )$ admits an expansion in the parameter $\epsilon$. 
For $\epsilon \ll 1$ one can Taylor expand the differentiable functions $D,E$ to find that 
\beq\label{s-expand}
\tilde s_t = s_t + \delta(\epsilon) \;,\; \; \text{with} \;\; \delta(\epsilon) =  \epsilon  \; \sum_I g^\sigma_I \frac{\partial D}{\partial g_I}\big|_{E(s_t)}  + \mathcal{O}(\epsilon^2)\, ,
\eeq
where $g^\sigma = \, \sigma_{a_t} \hat\ast E(s_t)$ and $g = E(s_t)$, and with the index $I=1,\dots,N$. The analog expression holds for $t+1$ i.e.\,$\tilde s_{t+1} = s_{t+1} +\mathcal{O}(\epsilon)$.
If $a_t = a_t(s_t)$ is differentiable function of the states\footnote{Although neural networks modeling the policies modeling the data-distribution contain non-differentiable components they are differentiable almost everywhere.} one may Taylor expand the Koopman operator using \Eqref{s-expand} as\footnote{The reader may wonder about the explicit use of the index $\delta_n(\epsilon)$ expressing the dimension of the state. it is necessary here although we have simply used $s_t$ as an implicit vector without explicitly specifying any components throughout the work.}
\bea\label{expand-Koop}
K(a_t(\tilde s_t))  &=& K(a_t(s_t)) + \sum_i\sum_n \delta_n(\epsilon) \, a'_{in}(s_t) \, \mathcal{K}_i \nonumber  \\[0,2 cm] 
  &=& K(a_t(s_t)) + \epsilon \, \Delta + \mathcal{O}(\epsilon^2) \;\; .
\eea
where  and $a'_{in}(s_t) = \tfrac{\partial a_t}{ \partial_{s_{t,n}}}|_{s_t} $ and $n=1,\dots, dim(s_t) $ and with
\beq
\Delta = \sum_i\sum_n\sum_I  g^\sigma_I \frac{\partial D_n}{\partial g_I}\big|_{E(s_t)} a'_{in}(s_t) \, \mathcal{K}_i    \;\;.
\eeq
However, by the definition of the symmetry map $a_t(\tilde s_t)=a_t(s_t)$, thus the Koopman operators in \Eqref{expand-Koop} match $\forall \epsilon$.
By using the assumption on the error one infers from \Eqref{st-proof}  that
\bea
\tilde s_{t+1}\hspace{-0.25 cm} &=& \hspace{-0.25 cm} D\Big((\mathds{1}+ \epsilon \sigma_{a_t} )\hat\ast  \mathcal{K}(a_t) E(s_{t}) \Big) \\[0,2 cm] \nonumber 
 &=&  \hspace{-0.25 cm} D\Big(\big(\mathcal{K}(a_t) + \epsilon_a  \, \mathds{1}\big)\,(\mathds{1}+ \epsilon \sigma_{a_t} )\hat\ast  E( s_{t}) \Big)\\[0,2 cm] \nonumber 
 &=& \hspace{-0.25 cm}  D\Big( \mathcal{K}(a_t)(\mathds{1}+ \epsilon \sigma_{a_t} )\hat\ast E( s_{t}) + \epsilon_a  E( s_{t})  +\mathcal{O}(\epsilon_a \epsilon) \Big)
\\[0,2 cm] \nonumber 
 &=&  \hspace{-0.25 cm}  D\Big(\mathcal{K}(a_t) E( \tilde s_t) + \epsilon_a  E( s_{t}) +\mathcal{O}(\epsilon_a \epsilon)\Big)\;\;,
\eea
One may Taylor expand $D$ by making use of $\epsilon_a \ll 1$ as
\bea\label{finalmain2}
\tilde s_{t+1} = D\big(\mathcal{K}(a_t) E( \tilde s_t) \big)   &+&  \epsilon_a  \sum_I E_I( s_t) \frac{\partial D}{\partial g_I}  \Big|_{ \mathcal{K}(a_t) E( \tilde s_t)} 
\nonumber \\[0,2 cm] 
 &+&\mathcal{O}(\epsilon_a \epsilon)+ \mathcal{O}(\epsilon_a^2) \;\;.
\eea
Thus under the assumption that the numerical error is given by a violation of the commutation relation one finds $\tilde s_{t+1} = \tilde F (\tilde s_t,a_t) +\mathcal{O}(\epsilon_a) $ where
\beq
 \epsilon_a \, \sum_I E_I( s_t) \frac{\partial D}{\partial g_I}  \Big|_{ \mathcal{K}(a_t) E( \tilde s_t)} \sim \mathcal{O}(\epsilon_a)\;\; .
\eeq
By comparison with \Eqref{s-expand} one infers that the $\epsilon$-expansion is a good approximation to the real dynamic if  $\epsilon_a \ll \epsilon$.
This concludes the proof of theorem \ref{thm:main2}.
Let us note that the proof  for  state shifts arising from a sum of symmetry transformations as
\bea\label{eq:gen-sym-shift}
\tilde s_t  &= & D\Big(\,\big(\mathds{1} \;+\;  \sum_{I=1}^N \epsilon_I \, \sigma^I_{a_t} \big)\hat\ast E(s_t) \Big) \; , 
 \\[0,2 cm] \nonumber
 \tilde s_{t+1} &= & D\Big(\,\big(\mathds{1} \;+\; \sum_{I=1}^N \epsilon_I \, \sigma^I_{a_t} \big) \hat\ast E(s_{t+1}) \Big) \;\; .
\eea
is completely analogous, and an analogous theorem holds. We choose the number of symmetries to run over the dimension of the Koopman latent space $N$ as it is relevant for the {\bf KFC++} setup.
\section{Implementation details}
\label{app:implementation}

{ \bf The Koopman forward model}
\label{sec:FwdModel}
The KFC algorithm requires pre-training of a Koopman forward model $\mathcal{F}: \mathcal{S}\times \mathcal{A} \to \mathcal{S}$ as\vspace{0,2 cm}
\beq
\mathcal{F}^{c 
 }(s_t,a_t)
    =\begin{cases}
       {\bf  c=0 \;\;\text{VAE}:} \;\;  D\big( E(s_t) \big) = s_t \vspace{0,2 cm}\\ 
       {\bf  c=1 \;\;\text{forward prediction}  :} \\
     D\Big(\, \big(\mathcal{K}_0 + \sum_{i=1}^m \mathcal{K}_i \, a_{t,i} \big) E(s_t) \, \Big) = s_{t+1} 
    \end{cases} 
\eeq
where both of $E$ and $D$  are approximated by Multi-Layer-Perceptrons (MLP's) and  the bilinear Koopman-space operator approximation are implemented by a single fully connected layer for $\mathcal{K}_{i=0,\dots,m}$, respectively.
The model is trained on batches of the offline data-set tuples $(s_{t+1},s_t,a_t)$ and optimized via  the loss-function
\beq
 \bar{l}_2(\mathcal{F}^{1}(s_t,a_t)\,, \, s_{t+1}) \; + \; \gamma_2 \,  \bar{l}_2(\mathcal{F}^{0}(s_t + s_\epsilon ,a_t)\, ,\, s_t + s_\epsilon )\;\;,
\eeq
where $s_\epsilon$ is a normally distributed random sate vector shift with zero mean and a standard deviation of $6 \cdot 10^{-2}$ where the $\bar{l}_2$-loss is the Huberloss.
For training the Koopman forward model in \Eqref{eq:Fwdmodel} we split the dataset in a randomly selected training and validation set with ratios $70\%/30\%$. We then loop over the dataset to generate the required symmetry transformations and augment the replay buffer dataset tuple $(s_t, s_{t+1},a_t,r_t, \sigma_{a_t} )$ according to \Eqref{eq:symoptions} as
\bea\label{eq:symoptions2}
&\text{\bf KFC}: & \;\;   \big(s_t, s_{t+1},a_t,r_t, \sigma_{a_t} \big) \\\nonumber
&\text{\bf KFC++}:& \;\; \big(s_t, s_{t+1},a_t,r_t, U(a_t),U(a_t)^{-1} \big)\;\;.
\eea
It is more economic to compute the symmetry transformation as a pre-processing step rather than at runtime of the RL algorithm.

Let us reemphasize a couple of points addressed already in the main text to make this section self-contained. 
Note that  we have built upon the implementation of CQL \citep{CQL}, which is based on SAC \citep{SAC}.
\textbf{We also use the same hyperparameters for all the experiments as presented in the CQL and S4RL papers 
for baselines.}

{\bf Koopman forward model:}
For the encoder as well as the decoder we choose a three layer MLP architecture, with hidden-dimensions $[512,512, N]$ and $[512,512, m]$, respectively. Where $N = 32$ is the Koopman latent space dimension and $m$ is the state space dimensions following our notation from section \ref{prelim}.
The activation functions are ReLU \citep{relu}. We use an ADAM optimizer with learning rate $3 \cdot 10^{-4}$.
We train the model for 75 epochs. 
 The batch size is chosen to be $256$. Lastly, we employ a random normally distributed state-augmentation for the VAE training i.e.\,$s \to s+s_\epsilon$ with $s_\epsilon\in \mathcal{N}(0,1 \cdot 10^{-2})$.

{\bf Hyper-parameters KFC:}
 We take over the hyper-parameter settings from the CQL paper \citep{CQL} except for the fact that we do not use automatic entropy tuning of the policy optimisation step \Eqref{eq:policy_improvement3} but instead a fixed value of $\alpha = 0.2 $. 
 The remaining hyper-parameters of the conservative Q-learning algorithm are as follows:   $ \gamma= 0.99$ and $\tau =5 \cdot 10^{-3}$ for the discount factor and target smoothing coefficient, respectively. Moreover, the policy learning rate is $1\cdot10^{-4} $ and the value function learning rate is $3\cdot10^{-4} $ for the ADAM optimizers.
 We enable Lagrange training for $\tilde \alpha$ with threshold of $10.0$ and learning rate of $3\cdot10^{-4} $  where the optimisation is done by an ADAM optimizer. The minimum Q-weight is set to $10.0$ and the number of random samples of the Q-minimizer is $10$.  For more specifics on these quantities  see \citep{CQL}.  The model architectures of the policy as well as the value function are three layer MLP's with hidden dimension $256$ and $ReLU$-activation.
 The batch size is chosen to be $256$. 
 Moreover, the algorithm performs behavioral cloning for the first $40k$ training-steps, i.e.\,time-steps in an online RL notation.

 The KFC specific choices are as follows: the split of random to Koopman-symmetry state shifts is $20/80$ i.e.\,$p_K = 80 \%$, whereas the option for the symmetry-type is either "Eigenspace" or "Sylvester" referring to case (I) and (II) of \Eqref{eq:symoptions}, respectively. 
Regarding the random variables in the symmetry transformation generation process, we choose $\epsilon_i \in  \mathcal{N}(0,1\cdot10^{-4}) $ for case (II)  and  $\epsilon \in  \mathcal{N}(0,5\cdot10^{-5}) $ for case (I) in \Eqref{eq:symoptions} after normalizing $\sigma_a$ by its mean. While for the random shift we use $\tilde\epsilon \in  \mathcal{N}(0,3\cdot10^{-3}) $ which is the hyper-parameter used in \citep{S4RL}.

{\bf Computational setup:}  We performed the empirical experiments on a system with PyTorch 1.9.0a \citep{pytorch}
The hardware was as follows: NVIDIA DGX-2 with 16 V100 GPUs and 96 cores of Intel(R) Xeon(R) Platinum 8168 CPUs and NVIDIA DGX-1 with 8 A100 GPUs with  80 cores of Intel(R) Xeon(R) E5-2698 v4 CPUs.
The models are trained on a single V100 or A100 GPU.

\section{Ablation study}
\label{sec:ablation}

\begin{table*}[h!]
    \centering
    \begin{tabular}{l | c c c c}
        \toprule
        \textbf{Task Name} & \textbf{S4RL-($\mathcal{N}$)} & \textbf{KFC} & \textbf{KFC++} & \textbf{KFC++-contact}  \\
        \midrule
        cheetah-random & \textbf{52.3} & 48.6 & 49.2 & 49.4 \\
        cheetah-medium & 48.8 & 55.9 & 59.1 & \textbf{61.4}\\
        cheetah-medium-replay & 51.4 & \textbf{58.1} & \textbf{58.9} & \textbf{59.3} \\
        cheetah-medium-expert & \textbf{79.0} & \textbf{79.9} & \textbf{79.8} & \textbf{79.8} \\
        \midrule
        hopper-random & \textbf{10.8} & \textbf{10.4} & \textbf{10.7} & \textbf{10.7}  \\
        hopper-medium & 78.9 & 90.6 & \textbf{94.2} & \textbf{95.0} \\
        hopper-medium-replay & 35.4 & \textbf{48.6} & \textbf{49.0} & \textbf{49.1}\\
        hopper-medium-expert & 113.5 & 121.0 & 125.5 & \textbf{129}\\
        \midrule
        walker-random & \textbf{24.9} & 19.1 & 17.6  & 18.3\\
        walker-medium & 93.6 & 102.1 & \textbf{108.0} & \textbf{108.3} \\
        walker-medium-replay & 30.3 & \textbf{48.0} & 46.1 & \textbf{48.5} \\
        walker-medium-expert & 112.2 & 114.0 & \textbf{115} & \textbf{118.1} \\
        \bottomrule
    \end{tabular}
    \caption{We study the effect of combining S4RL-$\mathcal{N}$ based augmentation
    training on ``contact'' event transitions
    (state transitions where the agent makes contact with a surface, read \ref{app:contact-events} for more details), and  KFC++ augmentation training on non-``contact'' events.
    We see that KFC++-contact performs similarly and in most cases, slightly better than
    the KFC++ baseline, which is expected.
    We experiment with the Open AI gym subset of the D4RL tasks and report the mean normalized returns over
    5 random seeds.}
    \label{tab:contact_events}
\end{table*}

\subsection{Training on non contact events}
\label{app:contact-events}
 
To test the theoretical limitations in a practical setting one interesting experiment is to use the S4RL-$\mathcal{N}$  augmentation scheme when the current state 
is a ``contact'' state and to choose KFC++ otherwise.
We define a ``contact'' state as one where the agent makes contact with another surface;
in practice we looked at the state dimensions which characterize the velocity of the agent
parts, and if there is a sign difference between $s_t$ and $s_{t+1}$, then contact with a
surface must have been made. The details regarding which dimension maps to agent limb velocities is available in the official
implementation of the Open AI Gym suite \citep{brockman2016openai}.

Note that the theoretical limitations are twofold, firstly the theorems \ref{thm:main} and \ref{thm:main2} only hold for dynamical systems with differentiable state-transitions; 
secondly, we employ a Bilinearisation Ansatz for the Koopman operator. Contact events lead to non-continuous state transitions and moreover would require incorporating for external forces in our Koopman description, thus both theoretical assumptions are practically violated in our main empirical evaluations of  KFC and KFC++. By performing this ablation study which is more suitable for our Ansatz, i.e.\,the KFC++ part is only trained on "differentiable" trajectories without contact events.
In other words, the Koopman description is unlikely to give good information regarding $s_{t+1}$ if there is 
a ``contact event'', we simply use S4RL-$\mathcal{N}$ when training the policies, which we expect to give slightly better performance than the KFC++ baseline.

We report the results in Table \ref{tab:contact_events}, where we perform results on the 
OpenAI Gym subset of the D4RL benchmark test suite.
We denote the proposed scheme as \textbf{KFC++-contact} and compare to both S4RL-$\mathcal{N}$ and 
KFC++.
We see that KFC++-contact performs comparably and slightly better in some instances, compared to 
KFC++. 
This is expected as contact events may be rare, or the data augmentation given by the KFC++ model
during contact events is pseudo-random and itself is normally distributed, which may suggest
that KFC++ behaves similarly to KFC++-contact.
However, we do not say this conclusively.

\subsection{Magnitude of KFC augmentation}
\label{sec:mag_st}
To further try to understand the effect of using KFC for guiding data 
augmentation, we measure the mean L2-norm of the strength of the
augmentation during training.
We measured the mean L2-norm between the original state and the augmented
state, and found the distance to be $2.6 \times 10^{-4}$.
For comparison, S4RL uses a fixed magnitude of $1 \times 10^{-4}$ \citep{S4RL}
for all experiments with S4RL-Adv.
Since S4RL-Adv does a gradient ascent step towards the direction with the highest
change, its reasonable for the magnitude of the step to be smaller in comparison.
An adversarial step too large may be detrimental to training as the step can be
too strong.
KFC minimizes the need for making such a hyperparameter search decision.

\subsection{Symmetries}
\label{sec:app_syms}
\paragraph{ Global translation symmetry CartPole:}
\begin{figure}[t!]
    \centering
    \includegraphics[width=224 pt]{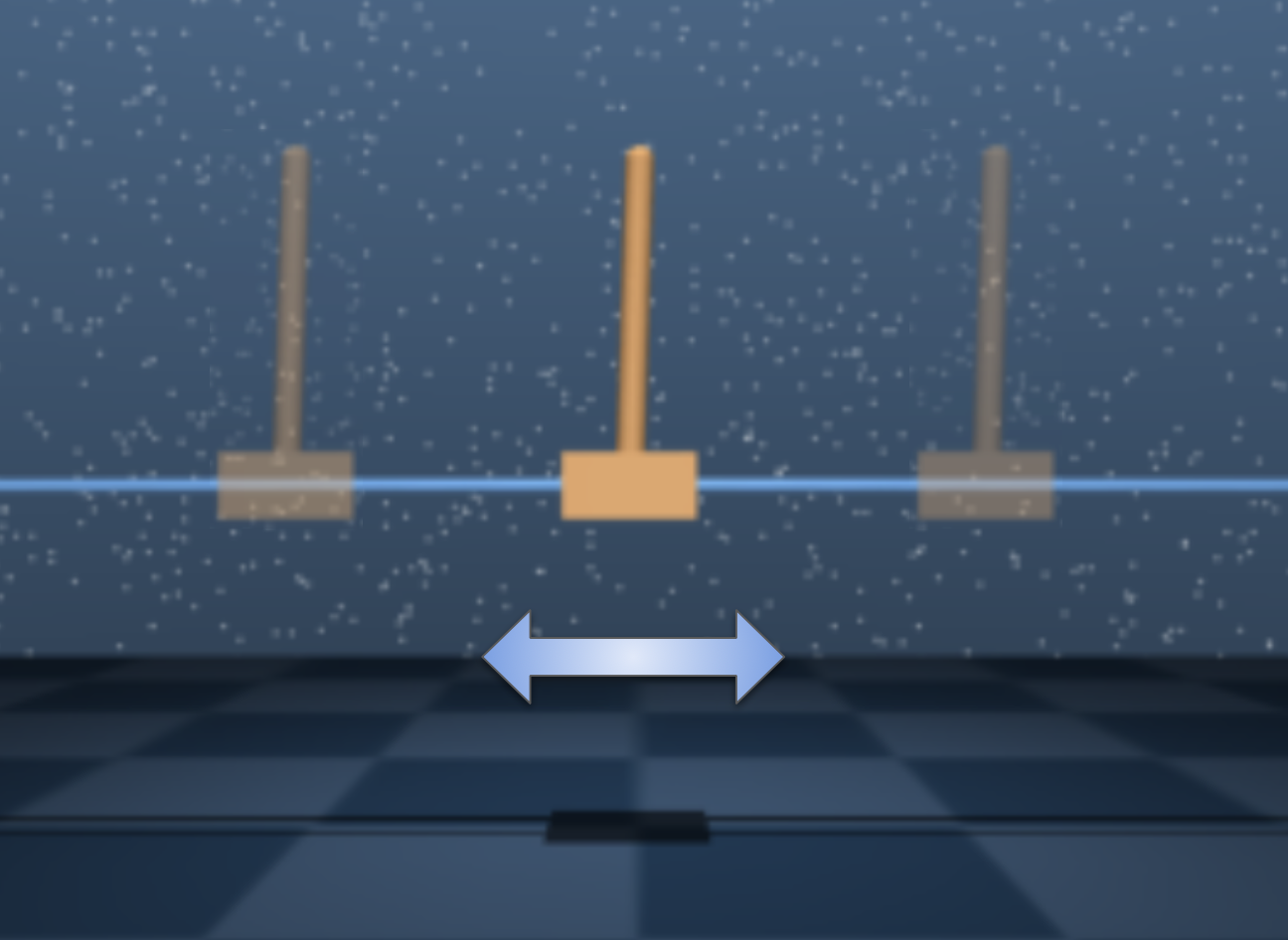}
    \caption{Depiction of translation symmetry of the  cartpole  environment in the Deepmind control suite \cite{tassa2020dmcontrol}. The dynamics is invariant of the position along the horizontal bar, i.e.\,under shifts to the left and right symbolized by the arrow.}
    \label{fig:cartpole}
    \vspace{-10pt}
\end{figure}
The cartpole's dynamics admits a translation symmetry w.r.t. to its position i.e.\,it is invariant under  left and right shifts, see fig.\ref{fig:cartpole}.
In the following we employ the KFC-algorithm for deriving  symmetries from the Koopman latent representation in the cartpole gym environment  \citep{brockman2016openai}. Moreover, we highlight that this is possible when purposefully employing imprecise  Koopman observables and operators.  We choose the Koopman observables to simply be the state vector $g = s =(x,\dot x, \theta, \dot\theta)$; position, angle and their velocities. The dynamics is not linear in these observables, thus the latter constitute a rough approximation. Nevertheless we will see that our approach is successful. 
{\bf Note that symmetries can be inferred from the approximate Koopman space representation}.

The trajectories are collected by a set of expert policies which admit the simple analytical form
\beq\label{eq:cartpolicy}
\pi(s,z) = \Theta\big(0.015 \,x \,+\, 0.066\,\dot x \,+\, 1.8\,\theta\,+\, 0.32\,\dot\theta + z \big)\;,
\eeq
where $\Theta$ is the Heaviside step function and $z \in \mathcal{U}(-0.2,0.2)$ a uniformly distributed random variable. The latter is sampled once at the beginning of each episode and is then fixed until the next one. This allows a for a wide variety of different expert policies from which we sample 100 trajectories each 1000 time-steps long.
Note that \Eqref{eq:cartpolicy} leads to an action set of $\{ 0,1\}$ as required by the cartpole gym environment \citep{brockman2016openai}.\footnote{ The physical action action set is $\{ -1,1\}$ which we use to train the Koopman model. Note that naturally a physical action space is required for our approach.} 

We empirically train a modified  Koopman forward model  \Eqref{eq:Fwdmodel} with trivial state embedding $g=s$ on the collected datasets and find that the approximate Koopman operator takes the form
\beq
\mathcal{K}\big(a=0\big) \;=\;
\begin{pmatrix}\label{eq:Kcart1}
1 & 0.02 &-0.01 &0.001\\
0&1 &0.6&0.8\\
0& 0& 1&0.02\\
0&-0.01&-0.7&-0.2\\
\end{pmatrix}
\eeq
\beq\label{eq:Kcart2}
\mathcal{K}\big(a=1\big) \;=\;
\begin{pmatrix}
1 & 0.02 &-0.009 &0\\
0&1 &0.6&0.8\\
0& 0& 1&0.02\\
0&-0.009&-0.7&-0.2\\
\end{pmatrix}
\eeq
 for the actions  $\{ 0,1\}$, respectively.\footnote{We have set numerical values at order $\mathcal{O}(10^{-4})$ and smaller to zero. Whiteout this truncation the subsequently derived relations will hold approximately instead of precisely, e.g.\,the vanishing commutation relation with the symmetry operation. Moreover, we have rounded the numbers to one significant digit.}

The translation symmetry  generation matrix emerges when imposing it to be commuting with the Koopman operator taking the form
\beq
\sigma_0 \;=\;
\begin{pmatrix}
-1& c_1&c_2&c_3\\
0&0 &0& 0\\
0&0 &0& 0\\
0&0 &0& 0\\
\end{pmatrix}
\eeq 
where $c_1=-2.35,c_2=-25.95,c_3=-2$ for $a=0$ and $c_1=-2.60759,c_2=-29.0296,c_3=-2.22222$ for $a=1$.

The translation  symmetry shift then takes the general form 
\beq\label{eq:cartsym}
\sigma_{a}^\epsilon = \mathds{1} + \epsilon \, \sigma_0 =
\left(\begin{array}{>{\columncolor{blue!20}}cccc}
    \rowcolor{red!20}
1-\epsilon& \epsilon\,c_1&\epsilon\,c_2&\epsilon\,c_3\\
0&1 &0& 0\\
0&0 &1& 0\\
0&0 &0& 1\\
 \end{array}\right)
\,.
\eeq
for $\epsilon \in \mathbb{R}$. Note that the Decoder and Encoder from and to Koopman space are simply the identify map in this case as we chose $g =s$.
 It is easy to see that \Eqref{eq:cartsym} indeed is a translation symmetry. Note that the blue-shaded area guarantees  that the position does not affect any of the other state variables. The latter, are mapped to themselves by  the identity matrix in the lower right corner. Moreover, the red-shaded area implies that the initial position is mapped to another position by the symmetry map, controlled by  the arbitrary $\epsilon$-parameter.
 We conclude that the translation symmetry operation emerges when imposing it to be commuting with the Koopman operator.

We show in theorem \ref{thm:main2} that for $\epsilon \ll 1$ \Eqref{eq:cartsym} is a local Lie symmetry.
However, in this case we may show that the local symmetry can be extended to any global value. We take the matrix exponential of the generator $\sigma_0$ and find that it is equivalent to 
\bea\label{eq:cartsymglobal}
\exp\big( \sigma_0 \, t\big) = \hspace{3.2cm}\nonumber \\[0.2cm]\nonumber
\left(\begin{array}{>{\columncolor{blue!20}}cccc}
    \rowcolor{red!20}
e^{-t}& c_1 \,(1-e^{-t})&c_2\,(1-e^{-t})&c_3\,(1-e^{-t})\\
0&1 &0& 0\\
0&0 &1& 0\\
0&0 &0& 1\\
 \end{array}\right) ,
\eea
for a parameter $t \in \mathbb{R}$. By comparison to \Eqref{eq:cartsym} one infers that
\beq
\exp\big( \sigma_0 \, t\big)  = \sigma_{a}^{\epsilon} \;\;\; \text{if} \;\;\; \epsilon=1-e^{-t} \;\;.
\eeq
In other words the composition of local symmetry transformations lead to extended "global" ones.

{\bf KFC++ algorithm:} The translation symmetry can easily be inferred  in a self supervised manner  algorithmically by the  KFC++ approach in \Eqref{eq:symoptions} as
\vspace{0.1 cm}
\bea
& {\bf\sigma^\epsilon_{a} \;\; :\;\; }& \,s \mapsto \tilde s =  \big(\mathds{1} +
\sigma_{0}(\vec\epsilon)\big) (s)   \, , \nonumber\\[0.2cm] \nonumber
&\sigma_{0}(\vec\epsilon)& =  {\bf \mathscr{R}e} \big(\, U(a) \, \text{diag}( \epsilon_1, \epsilon_2, \epsilon_3,  \epsilon_4) \, U^{-1}(a)  \big)\;,\;\;
\vspace{0.1 cm}
\eea
for $a=0,1$.
One computes the eigenvalues $\lambda_i,\, i=1,..,4$ and vectors $U(a=0,1)$ of the Koopman operators \Eqref{eq:Kcart1} and \Eqref{eq:Kcart2}. Notable the systems admit an eigenvector $U_0 = (1,0,0,0) $ to the eigenvalue $\lambda_0 = 1$. One infers that
\bea\label{eq:cartKFC}
\sigma^{\epsilon_1}_{a} = \mathds{1} + \sigma_{0}(\epsilon_1, \epsilon_2=0, \epsilon_3=0,  \epsilon_4=0) = \nonumber \\ \nonumber
\left(\begin{array}{>{\columncolor{blue!20}}cccc}
    \rowcolor{red!20}
1+\epsilon_1& -\epsilon_1\,c_1&-\epsilon_1\,c_2&-\epsilon_1\,c_3\\
0&1 &0& 0\\
0&0 &1& 0\\
0&0 &0& 1\\
 \end{array}\right) \;\; .
\eea
Thus the translation symmetry is obtained by symmetry shifts w.r.t.\,the first eigenvalue.
Moreover, we note that for the general setting i.e.\,
$\epsilon_i \neq 0 ,\, i=2,3,4$ one finds that the translation symmetry is naturally composed by other local symmetries as
\bea\label{eq:cartKFC2}
\sigma^{\vec\epsilon}_{a} = \mathds{1} + \sigma_{0}(\vec\epsilon) = 
\left(\begin{array}{>{\columncolor{blue!20}}cccc}
    \rowcolor{red!20}
1+\epsilon_1& \cdot & \cdot& \cdot\\
0&  \cdot &  \cdot &  \cdot \\
0&\cdot &  \cdot &  \cdot\\
0&\cdot &  \cdot &  \cdot\\
 \end{array}\right) \;\; ,
\eea
where we have used the symbol "$\cdot$" as proxy for in generally non-zero elements.

\begin{figure*}[t!]
    \centering

    
    \begin{subcaption}{Hopper position state shifts}
    \includegraphics[width=\textwidth/3]{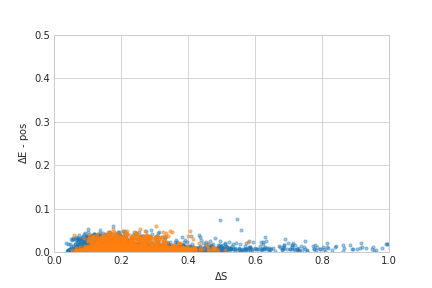}
     \includegraphics[width=\textwidth/3]{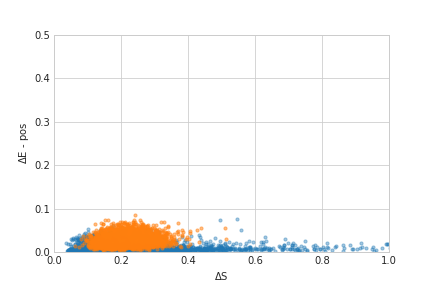}
    \end{subcaption}
    ~
    \begin{subcaption}{Hopper velocity state shifts}
    \includegraphics[width=\textwidth/3]{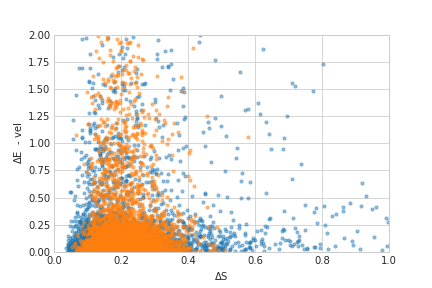}
      \includegraphics[width=\textwidth/3]{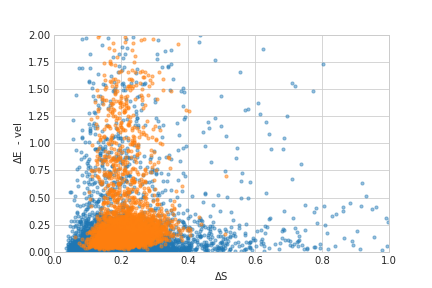}
    \end{subcaption}
         \includegraphics[width=\textwidth/3]{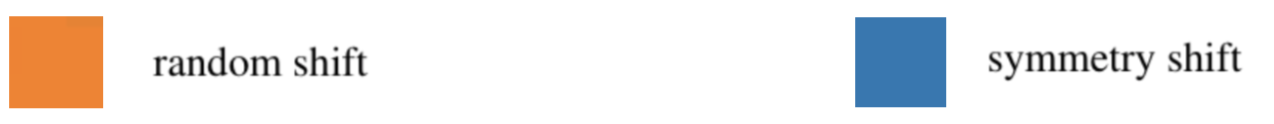}
     \caption{Symmetry shifts vs. random latent space shifts of same magnitude in the mean of $\Delta S$ compared by its accuracy in the online evaluation in the gym Hopper-v3 environment. The first line shows the result where $s_t$ and $s_{t+1}$  are shifted by the same N-dimensional random variable. For the ablation study in line two we sample twice from the random variable for the shifts $s_t$ of $s_{t+1}$, respectively.   }
    \label{fig:sym-exps}
\end{figure*}

\paragraph{ Local precision symmetry shifts Mujoco:}
In this section we provide a quantitative analysis of the symmetry shifts generated by KFC++ in comparison to  random shifts of the latent states. For the latter we choose the variance such that absolute state shift 
\beq
\Delta S := |\tilde s_t -s_t  | +  |\tilde s_{t+1} -s_{t+1}|
\eeq
averaged over all sampled data points (D4RL) is comparable between the two setups. Moreover we simply use the encoder
We then compare how well the shifted states are in alignment with the actual dynamic using the online Mujoco environment .\footnote{Note that this requires a minor modification of the environment code by a function which allows to set states to specific values.} To do so we set the environment state to the value $\tilde s_t$ and use the action $a_t$ to perform the environment step which we denote by $s_{t+1} = M(s_t,a_t)$\footnote{$M(s_t,a_t)$ denotes the step done by the active gym environment.} The performance metric is then given by
\beq\label{eq:deltaE}
\Delta E := |s_{t+1} - M(\tilde s_t,a_t)| \;\;.
\eeq
Note that $\Delta E$ is simply the error to the actual dynamic of the environment.
We take the model trained on the hopper-medium-expert dataset from section appendix \ref{app:contact-events} to compute the symmetry shift.
Moreover, for the comparison in \Eqref{eq:deltaE} we split  the state space into positions and velocities of the state vector, i.e.\,the first five elements of $s_{t+1} - M(\tilde s_t,a_t)$ give position while the remaining ones given velocities. The norm is taken subsequently.
 For the results see figure \ref{fig:sym-exps}.
 
 We conclude that there is a qualitative difference between the distributions obtained. Most notable the symmetry shift allows for a controlled shift by a larger $\Delta S$ while still maintaining a small error $\Delta E$. This amounts to a wide but accurate exploration of the environments state-space. Although we cannot say this conclusively we expect  this to be the reason for the performance gains of KFC++ over random shifts.

\subsection{VAE without Koopman Operator}

\begin{table}[h!]
    \centering
    \begin{tabular}{l | c c c c}
        \toprule
        \textbf{Task Name} & \textbf{S4RL-(Adv)} & \textbf{KFC} & \textbf{KFC++} & \textbf{VAE}  \\
        \midrule
        Gym  & 61.61 & 66.36 & \textbf{67.76} & 52.90 \\
        %
        Antmaze  & 62.88 & 65.52 & \textbf{67.35} & 54.26 \\
        \bottomrule
    \end{tabular}
    \caption{Results CQL with data-augmentation by a VAE with additive random latent space shifts.  The results are averaged on the Gym and Antmaze tasks, respectively. We report the best results of the VAE over a hyperparameter search of the variance of the random variable. }
    \label{tab:ablationVAE}
\end{table}
We have performed an additional ablation study to compare KFC to a conventional VAE, see Table \ref{tab:ablationVAE} for the results over the 
OpenAI Gym and Antmaze subset of the D4RL datasets. 
We find increased performance of KFC/KFC++ over data-augmentation by a common VAE with random latent space shifts.
This ablation study is similar to the VAE baseline used in S4RL \cite{S4RL}, but the main difference is that we replace the Koopman operator in KFC
with adding random noise in the learned latent representation of the VAE.
We add noise of the form of $\mathcal{N}(0,\epsilon \times \mathcal{I})$; we perform a hyperparameter sweep over values for $\epsilon$ between $10^{-1}$ and $10^{-5}$. 
We report results with the best-performing noise  value ($\epsilon = 3 \times 10^{-3}$).
The VAE does not improve the performance over the S4RL baseline or KFC variants.
We conclude that  KFC/KFC++ is still the best-performing method due to the use of symmetry shifts and its Koopman latent representation, which is 
significantly better than adding random noise to the latent representation.

\subsection{Prediction Model}
As a further ablation study it is of interest to compare the symmetry induced state shifts tuple $(\tilde s_t, \tilde s_{t+1})$ to one  obtained by the forward prediction of our VAE-forward model \Eqref{eq:Fwdmodel}. 
\newline
{\bf KFC++prediction:}
In particular we use the forward prediction on $\tilde s_t $ obtained by the KFC++ shift, i.e.\, $\tilde s_{t+1} = \mathcal{F}^{c=1}(\tilde s_t,a_t)$. 
 We refer to this setup in the following as KFC++prediction. See table \ref{tab:prediction-model} for the results. 
 \newline
 {\bf Fwd-prediction:}
 Moreover, we study the case where we  $\tilde s_t $ is obtained by a shift with a normally distributed random variable $\mathcal{N}(0,6\cdot 10^{-3})$ and then simply forward predict as $\tilde s_{t+1} = \mathcal{F}^{c=1}(\tilde s_t,a_t)$. This is comparable to conventional model based approaches employing our simple VAE forward model. This constitutes an ideal systematic comparison as the model, hyper-parameters and training procedure are identical to the one used in the KFC variants. Moreover, the variance of the random variable is such that the distance to the augmented states is comparable to the ones in KFC, see appendix \ref{sec:mag_st}.  See table \ref{tab:prediction-model2} for the results. 
\begin{table*}[h!]
    \centering
    \begin{tabular}{l | c c c c}
        \toprule
        \textbf{Task Name} & \textbf{KFC} & \textbf{KFC++} & \textbf{KFC++-prediction}  \\
        \midrule
        cheetah-random  & 48.6 & \textbf{49.2} & 46.5 \\
        cheetah-medium & 55.9 &\textbf{59.1} & 53.7\\
        cheetah-medium-replay & \textbf{58.1} & \textbf{58.9} & 55.3 \\
        cheetah-medium-expert & \textbf{79.9} & \textbf{79.8} & 76.3 \\
        \midrule
        hopper-random & \textbf{10.4} & \textbf{10.7} &  \textbf{10.8} \\
        hopper-medium & 90.6 & \textbf{94.2} & 90.5 \\
        hopper-medium-replay & \textbf{48.6} & \textbf{49.0} & 44.2 \\
        hopper-medium-expert & 121.0 & \textbf{125.5} & 121.2\\
        \midrule
        walker-random & \textbf{19.1} & 17.6  & 15.6 \\
        walker-medium & 102.1 & \textbf{108.0} & 105.3 \\
        walker-medium-replay & \textbf{48.0} & 46.1 & 45.2 \\
        walker-medium-expert & 114.0 & \textbf{115.6} & 114.5 \\
        \bottomrule
    \end{tabular}
    \caption{Results with prediction model  KFC++-prediction on the Open AI Gym subset of the D4RL tasks. We report the mean normalized episodic rewards over 5 random seeds similar to the original D4RL paper \cite{d4rl}.}
    \label{tab:prediction-model}
\end{table*}

\begin{table*}[h!]
    \centering
    \begin{tabular}{l|l | c c |  c }
        \toprule
         
        \textbf{Domain} & \textbf{Task Name} &  \textbf{KFC} & \textbf{KFC++} & \textbf{Fwd-prediction}  \\
        \midrule
        \midrule
        \multirow{6}{*}{AntMaze} 
        & antmaze-umaze & \textbf{96.9} & \textbf{99.8} & 92.7 \\
        & antmaze-umaze-diverse & \textbf{91.2}& \textbf{91.1}  & 90.1 \\
        & antmaze-medium-play  & 60.0 & \textbf{63.1}  & 60.8\\
        & antmaze-medium-diverse  & 87.1& \textbf{90.5}  & 88.0\\
        & antmaze-large-play &  24.8 & \textbf{25.6}  & 23.1\\
        & antmaze-large-diverse  & \textbf{33.1} & \textbf{34.0} & 29.3 \\
        \midrule
        \midrule
        \multirow{12}{*}{Gym} 
        & cheetah-random  & 48.6 & 49.2  & \textbf{50} \\
        & cheetah-medium  & 55.9 & \textbf{59.1} & 50.1 \\
        & cheetah-medium-replay   & \textbf{58.1} & \textbf{58.9}  &  56.4 \\
        & cheetah-medium-expert  & \textbf{79.9} & \textbf{79.8} & 71.3  \\
        & hopper-random &  \textbf{10.4} & \textbf{10.7}  & \textbf{10.4}  \\
        & hopper-medium  & 90.6 & \textbf{94.2} & 82.3  \\
        & hopper-medium-replay  & \textbf{48.6} & \textbf{49.0}& 40.8  \\
        & hopper-medium-expert  & 121.0 & \textbf{125.5} & 120.3 \\
        & walker-random &  \textbf{19.1} & 17.6 & 18.4 \\
        & walker-medium  & 102.1 & \textbf{108.0} & 103.2 \\
        & walker-medium-replay & \textbf{48.0} & 46.1 & 41.7 \\
        & walker-medium-expert  & 114.0 & \textbf{115.3} & 111.8 \\
        \midrule
        \midrule
        \multirow{8}{*}{Adroit} 
        & pen-human  & \textbf{61.3} & 60.0 & 49.4 \\
        & pen-cloned & \textbf{71.3} & 68.4& 50.2 \\
        & hammer-human & 7.0 & \textbf{9.4}& 6.1  \\
        & hammer-cloned  & 3.0 & \textbf{4.2}& 4.2 \\
        & door-human   & 44.1 & \textbf{46.1}& 41.8  \\
        & door-cloned   & 3.6 & \textbf{5.6} & 1.2 \\
        & relocate-human  & \textbf{0.2} & \textbf{0.2}& 0.2  \\
        & relocate-cloned    & \textbf{-0.1}& \textbf{-0.1} & \textbf{-0.1}  \\
        \midrule
        \midrule
        \multirow{2}{*}{Franka} 
        & kitchen-complete & \textbf{94.1} & \textbf{94.9} & 90.0  \\
        & kitchen-partial  & 92.3 & \textbf{95.9} & 84.6 \\
        \bottomrule
    \end{tabular}%
    \caption{Results with prediction model Fwd-prediction on the D4RL tasks. We report the mean normalized episodic rewards over 5 random seeds similar to the original D4RL paper \cite{d4rl}.}
    \label{tab:prediction-model2}
\end{table*}

We conclude that {\bf KFC++prediction} falls behind both {\bf KFC} as well as {\bf KFC++}. This was expected as the symmetry shift in the latter alters the original data points by means of the VAE which admits not only a much higher accuracy but also advanced generalisation capabilities to out of distribution values. 
\newline
More interesting however is  that {\bf Fwd-prediction} falls behind both {\bf KFC} as well as {\bf KFC++} as well as {\bf KFC++prediction}. This is a strong indicator that the symmetry shifts provide superior out-of-distribution data for training a Q-learning algorithm.

\subsection{Discussion}

Although our KFC framework is suited best for the description of environments described by control dynamical systems the  learned (in self-supervised manner) linear Koopman latent space representation may be applicable to a much wider set of tasks.
However, there are notable shortcomings to the current implementation both conceptually as well as practically. The bilinearisation in \Eqref{eq:Fwdmodel} of the latent space theoretically assumes the dynamical system to be governed by \Eqref{eq:BilinKoop}, which is rather restrictive. Although a general non-affine control system admits a bilinearisation \citep{brunton2021modern} it  generically requires the observables to depend on the action-space variables implicitly.
Secondly, the Koopman operator formalism is theoretically defined by its action on an infinite dimensional observable space. The  finite-dimensional approximation i.e. the latent space representation of the Koopman forward model in \Eqref{eq:Fwdmodel}  lacks accuracy due to that.
On the practical side our formalism requires data pre-processing which is computationally expensive, i.e. solving the Sylvester or eigenvalue problem for every data-point.
Moreover, the Koopman forward model in \Eqref{eq:Fwdmodel} serves as a function approximation to two distinct tasks. Thus one faces a twofold accuracy vs. over-estimation problem which needs to be balanced.
The systematic error in the VAE directly imprints itself on the state-data shift in Eqs.~(\ref{eq:symshift}) and (\ref{eq:symoptions}) and may thus conceal any potential benefit of the symmetry considerations. Lastly, the dynamical symmetries do not infer the reward. Thus the underlying  working assumption is that the reward should not vary much in \Eqref{eq:symshift}.

\paragraph{A note on the simplicity of the current algorithm:}Let us stress a crucial point. Algorithmically the symmetry maps are derived in two distinct ways {\bf KFC} and {\bf KFC++}. The latter, constitute a simple starting point to extract symmetries from our setup. More elaborate studies employing the extended literate on Koopman spectral analysis are desirable. Moreover, it is desirable to extend our framework to more complex latent space descriptions.  It is our opinion that by doing so there is significant room for improvement both in the accuracy of the derived symmetry transformations as well their induced  performance gains of Q-learning algorithms.
 Note that currently our VAE model is of very simple nature and the symmetries are extracted in a rather uneducated way. While the Sylvester algorithm simply converges to one out of many symmetry transformations  for the   KFC++ algorithm we omit all the information of the imaginary part, let alone utilize concrete spectral information.

\end{document}